\pdfoutput=1

\documentclass[11pt]{article}

\usepackage[final]{acl}

\usepackage{times}
\usepackage{latexsym}

\usepackage[T1]{fontenc}

\usepackage[utf8]{inputenc}

\usepackage{microtype}

\usepackage{inconsolata}

\usepackage{graphicx}
\usepackage{enumitem}

\usepackage{hyperref}
\usepackage{xcolor}
\usepackage{booktabs}
\usepackage{longtable}
\usepackage{tabularx}
\usepackage{amsmath}
\usepackage{multirow}
\usepackage{graphicx}
\usepackage{colortbl}
\usepackage{graphicx}
\usepackage{amsmath} 
\usepackage{array} 

\hypersetup{
    colorlinks=true,
    linkcolor=blue,
    filecolor=magenta,      
    urlcolor=cyan,
}
\usepackage[T1]{fontenc}

\usepackage[utf8]{inputenc}
\usepackage{multicol}

\usepackage{microtype}
\title{A Character-Centric Creative Story Generation via Imagination}

\author{
 \textbf{Kyeongman Park\textsuperscript{1}},
 \textbf{Minbeom Kim\textsuperscript{1}},
 \textbf{Kyomin Jung\textsuperscript{1}}
\\
 \textsuperscript{1}Seoul National University
\\
    \texttt{zzangmane@snu.ac.kr}, \texttt{minbeomkim@snu.ac.kr},
    \texttt{kjung@snu.ac.kr}
}

\begin{document}
\maketitle
\begin{abstract}
Creative story generation has long been a goal in NLP research. 
Although existing methods can produce lengthy and coherent narratives, they often lack human creativity in terms of diversity and depth of character.
To address this, we introduce a novel story generation framework called \textbf{CCI} (\textbf{C}haracter-centric \textbf{C}reative story generation via \textbf{I}magination). CCI features two modules for creative story generation: IG (Image-Guided Imagination) and MW (Multi-Writer model).
In the IG module, we utilize a text-to-image model to create visual representations of key story elements, such as characters, backgrounds, and main plots, in a more novel and concrete manner than text-only approaches.
The MW module uses these story elements to generate multiple persona-description candidates and selects the best one to insert into the story, thereby enhancing the richness and depth of the narrative.
We compared the stories generated by CCI and baseline models through statistical analysis, as well as human and LLM evaluations. The results showed that the IG and MW modules significantly improve various dimensions of creative storytelling. Moreover, our framework supports interactive multi-modal story generation with users, opening up new possibilities for human-LLM integration in cultural development.\footnote{Project page : \url{https://www.2024cci.p-e.kr/}}.

\end{abstract}
\section{Introduction}

\begin{quote}
 \begin{minipage}{1.0\linewidth}
\centering
\textit{``The lunatic, the lover, and the poet,\\
Are of imagination all compact.''\\}

\end{minipage}
\end{quote}

\noindent

Great literary works are products of imagination \cite{camargo2017creativity}. When writing a story, human writers often begin by imagining creative elements—such as unique characters, dramatic climaxes, or magnificent backgrounds—and then build the narrative around them. Even as the writing progresses, their creative imagination remains active, rendering character descriptions vivid, as if truly brought to life before the reader.

\begin{figure}[t]
\centering
\includegraphics[width=0.5\textwidth]{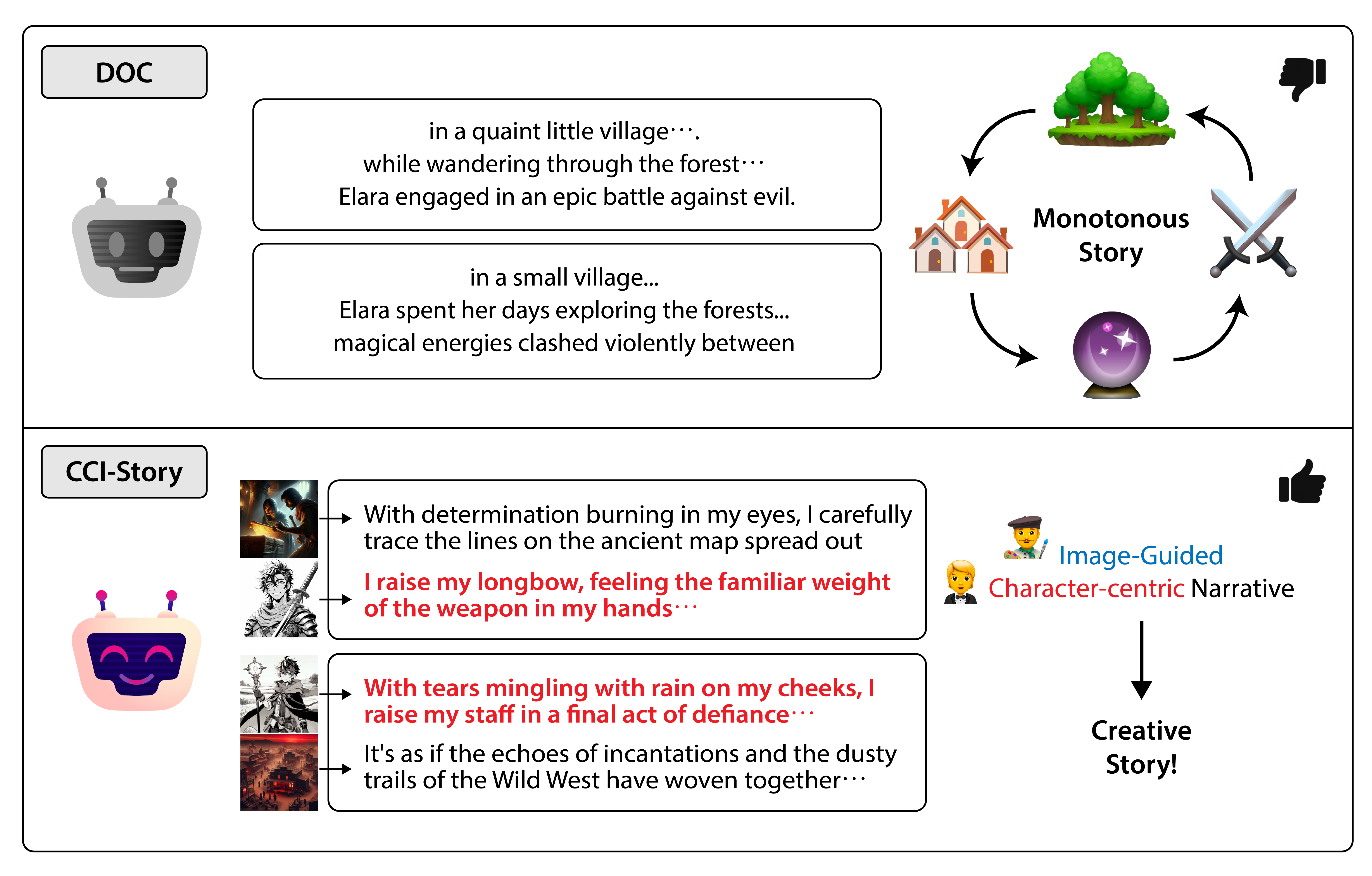}
\caption{
Comparison between DOC and Our CCI-Story Approach. DOC generates stories on similar topics in a monotonous manner. In contrast, our work leverages images to create stories that are not only diverse and creative in their themes but also richer in content, centered around the persona of the main character. 
}
\vspace{-5mm}
\end{figure}

By contrast, in our survey analysis (Appendix \ref{sec:survey}), previous state-of-the-art story generation models have two critical limitations in creativity : (1) insufficient diversity and detail in generating key story elements (e.g., characters, settings, and plots), and (2) an inability to construct complex and nuanced protagonists.

Despite significant advancements in automatic story generation tasks \cite{rashkin2020plotmachines,tang2022etrica,fan2018hierarchical}, such systems have largely prioritized coherence and length~\cite{yang2022re3,yang2022doc,park2024longstory,wen2023grove,achiam2023gpt,touvron2023llama} and often overlooked the importance of creative story elements and the deep psychology of characters.


To address these issues, we propose a novel framework for creative story generation called CCI (\textbf{C}haracter-centric \textbf{C}reative Story generation via \textbf{I}magination). CCI employs three main components: IG (Image-Guided Imagination) to create diverse and concrete story elements using visual representations, MW (Multi-Writer) to enrich the complexity and specificity of the protagonist’s character, and a Specification process that interconnects IG and MW by expanding on character and main plot information.

First, IG leverages AI-generated images to ensure diversity and detail in story elements, including characters, backgrounds, and climactic scenes. Specifically, IG produces visual representations through a text-to-image model and then inputs them into an LVLM (Large Vision Language Model), such as GPT-4o, to extract detailed textual descriptions. The variety of AI-generated images reduces repetition across samples, and visual details (color, texture,  etc.) enhance specificity, thereby increasing overall creativity in the story elements.

Next, the Specification process integrates the IG and MW steps. It enriches character descriptions with detailed aspects—such as dark secrets, habits, and speech tones—collectively referred to as a persona \citep{park2024enhancing}. It also refines the main plot details through the Chain of Ask Why method \citep{wen2023grove}.

Finally, MW generates multiple candidates for specific protagonist descriptions to ensure that each paragraph contains coherent and richly developed character traits. These candidates are then filtered for relevance to the established persona and their similarity to previous sentences. The candidate with the best continuation score is incorporated into the story, deepening character complexity and contributing to a more creative narrative.

Our statistical analysis demonstrates that stories generated by CCI consistently achieve greater diversity and stronger connections to the protagonist’s persona. Human and LLM evaluations show a 79.5\% average preference for our framework over existing baselines. Ablation studies further validate that IG enhances novelty, concreteness, and overall story quality, while MW improves character coherence, vividness, and the effective use of story elements.

Additionally, our framework can be applied to multi-modal interactive story generation. Human and LLM annotators reported that CCI better expresses their custom images, showing a 58.95\% average preference over \texttt{w/oMW}.

In summary, our main contribution is CCI, a framework for creative story generation that produces imaginative narratives and diverse story elements by actively utilizing visual representations (IG) and incorporating multiple character personas (MW) to enrich the protagonist’s portrayal.



\section{Related Works}
\subsection{Creative Story Generation}

Previous story generation models have primarily focused on ensuring coherence rather than enhancing creativity. While they have shown their coherent results do not sacrifice diversity by measuring the repetitiveness within each generated story, this approach falls short in addressing creativity directly \citep{rashkin2020plotmachines,tang2022etrica,park2024longstory}. Efforts to increase diversity have generally involved training on large datasets, but extending creativity beyond the scope of the training data remains a challenge \citep{fan2018hierarchical,rashkin2020plotmachines,tang2022etrica,park2024longstory}. Recently, the DOC \citep{yang2022re3,yang2022doc} utilizing large language models (LLMs) \citep{achiam2023gpt,touvron2023llama,jiang2023mistral} for generating longer stories, has increased randomness in sampling story elements by setting a high temperature. However, this method still falls short of achieving creative storytelling. For these limitations, CRITICS \citep{bae2024collective} collaborate with humans to leverage creativity, but our approach aims to stimulate the AI's own creativity, without human's intervention, through a fully automatic story generation framework.


\subsection{Multi-modal Story Generation}


Visual storytelling \citep{shen2023storygpt,gong2023talecrafter,huang2016visual,yang2023attractive,wong2022compute,huang2019hierarchically,pan2024synthesizing,rahman2023make,li2019storygan,jeong2023zero,cheng2024theatergen,maharana2022storydall,li2018video} has gained significant attention recently, thanks to advancements in latent diffusion models \citep{rombach2022high}. In this paper, we also target multimodal story generation \citep{zhou2023multimodal,lu2022imagination,guo2022visually,koh2023grounding,ge2023making}, which involves creating a story based on various multimodal features, with a particular focus on images in our work. Many previous approaches \citep{min2021deep,lovenia2022every,wang2024sco,shuster2019engaging,xu2021imagine} typically generate very short descriptions (about 0.2k tokens) for each image. Recently, SEED-Story \citep{yang2024seed} has succeeded in generating up to 25 consecutive scene images and creating relatively long stories for each image using Multimodal LLMs \citep{betker2023improving}. However, unlike SEED-Story, our work focuses more on the quality of natural language, resulting in much longer, more diverse, detailed, and human-like stories. Additionally, we do not require any fine-tuning of LLMs, which significantly reduces training costs.




\subsection{Character-centric Neural Story Generation}

Creative literature involves not only diverse but also detailed descriptions. Particularly, incorporating contradictory and complex personalities into characters is a hallmark of great literary works \cite{camargo2017creativity}. There have been many attempts to impart human personalities to language models \citep{jiang2023personallm,mao2023editing,lyons1972human,weng2024controllm}. The personal traits generally assigned to language models include preferences \citep{he2024afspp,serapio2023personality}, role \citep{liu2020character}, personality \citep{jiang2023personallm,mao2023editing,weng2024controllm,akoury2020storium,zhang2022persona,jiang2024evaluating}, knowledge \citep{lyons1972human,park2024enhancing}, habits \citep{gurung2024chiron,cavazza2005dialogue,park2023generative}, appearance \citep{gurung2024chiron,cheng2024theatergen}, psychology \citep{gurung2024chiron,eger2017character,xu2020controllable}, and emotions \citep{brahman2020modeling,xie2022psychology}. Such approaches result in more vivid and natural outputs \citep{liu2020character,eger2017character,fay2014driving,zhang2022persona,brahman2020modeling}. In the story generation task, the model should consistently maintain character's characteristics to ensure coherence \citep{he2024afspp,eger2017character}. In our work, we actively create and maintain diverse and complex character identity, called persona \citep{park2024enhancing,zhang2022persona,jiang2023personallm}. IG is responsible for creating these personas, while MW utilizes them. While writing long stories, we also update the character's persona according to the context \citep{he2024afspp,xie2022psychology}.


%
\subsection{Human-Interactive Story Generation}

 Many story generation frameworks provide methods for collaborative interaction with humans \citep{radwan2024sard,yang2022re3,yang2022doc,gong2023talecrafter,goldfarb2019plan,lee2022coauthor}. This approach allows users to exert more precise and detailed control over the stories. These interactions have been mainly passage-by-passage feedback in the past \citep{gong2023talecrafter,goldfarb2019plan,lee2022coauthor}, but recent state-of-the-art works \citep{radwan2024sard,yang2022re3,yang2022doc} have suggested a method of controlling the high-level outline through interaction. However, no work has supported multimodal interaction with users. Our methodology supports multimodal interaction with users during the high-level planning phase.

\begin{figure*}
\centering
\includegraphics[width=1.0\textwidth]{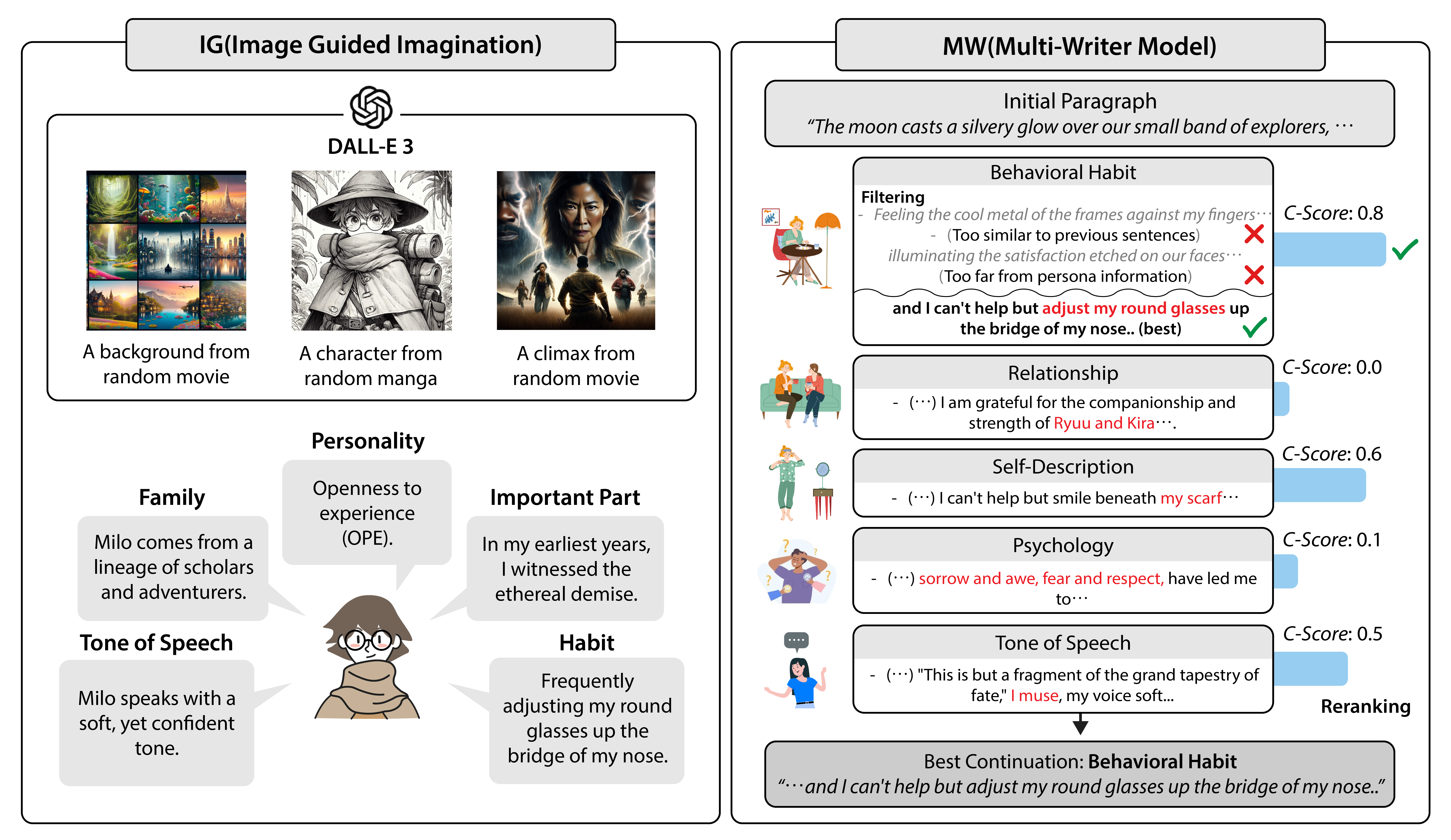}
\caption{
Our two main modules, IG and MW. In the IG module, we generate story elements using DALL-E 3 and then create the protagonist's persona through an iterative question-and-answer process. In the next step, the MW module generates multiple candidate descriptions for each ending of the initial paragraph, ensuring they reflect the protagonist's specific persona attributes. During this process, MW filters out candidates that are either too similar to previous sentences or deviate too far from the persona information (filtering examples are in the \texttt{Behavioral Habit} box). Finally, MW ranks the filtered candidates and selects the best one to continue the initial paragraph, using the \texttt{CS} (referred to as \texttt{C-Score} in the figure).
}
\vspace{-5mm}
\end{figure*}

\section{CCI Story Generation}

We utilize the DOC framework\cite{yang2022doc} as our backbone, aiming to address and significantly enhance the aspect of `creativeness' — an issue not extensively explored in the original work. To this end, we introduce two novel modules: Image-guided Imagination (IG) and Multi-writer (MW). 

\subsection{Image-guided Imagination (IG)}

To generate story elements that are diverse, novel, and specific, we propose \textbf{Image-guided Imagination (IG)}. IG consists of two steps: (1) Visual Imagination and (2) Image-guided story elements creation.

\subsubsection{Visual Imagination}\label{Visual Imagination} The IG module first generates background, protagonist, and main plot images from fixed short prompts. The text-to-image model, DALL-E 3\citep{betker2023improving}, enhances the original prompt into a more detailed and diverse one using a recaptioning system.  DALL-E 3 then generates visual imagination based on the improved prompt. To leverage maximum imaginative potential for the DALL-E 3 model, we experiment with three candidate fixed short prompts, detailed in appendix \ref{sec:appendixE}. Additionally, in Section \ref{Diversity of Story Elements.}, we experiment with replacing DALL-E 3 with two Stable Diffusion-based models to verify that any visual imagination method can enhance creativity compared to text-only models.

\subsubsection{Image-guided Story Elements Creation}\label{Image-guided story elements creation} Using the images generated in the previous step, GPT-4o creates detailed descriptions of the images. Due to the unique details provided by Visual Imagination process, such as color, texture, and spatial information, GPT-4o can enhance creativity and diversity in the story elements creation. Example prompts are provided in Appendix \ref{sec:appendixa_1}.

\subsection{Specification}
\label{specification}

This stage further specifies the characters and main plot elements generated by IG.

\subsubsection{Character Specification} Given the character description and the image, GPT-4o generates the following persona information:
\begin{enumerate}
\item Dark secret.
\item Family environment.
\item Description his or her appearance oneself.
\item Specific way of speakings(tone of speech).
\item Personality.\citep{goldberg2013alternative},
\item The most significant events in life. 
\item Habitual behaviors.
\end{enumerate}

We use this persona information in a following step, the MW. Example prompts can be found in Appendix \ref{sec:appendixa_2}.

\subsubsection{Main Plot Specification} To resolve ambiguous points in the main plot, we employ the Chain of Asking Why \citep{wen2023grove} and Main Plot Specification. In the Chain of Asking Why, we instruct the LLM to identify N ambiguous points and generate answers for them. This process is repeated up to M times, incorporating previous answers into the original input. The process ends when the LLM either finds no further ambiguities or after M repetitions. With the clarified main plot, we provide all the previously generated information and ask the following questions:
\begin{enumerate}
\item Why this event must happen to the protagonist, in this background? 
\item How does the protagonist appropriately respond to this event, in this background?
\item Summarize all the answers provided above in five sentences. 
\end{enumerate}
This process not only reduces the imaginative burden in subsequent steps but also tightly integrates the main plot, background, and protagonist. Example prompts are provided in Appendix \ref{sec:appendixa_2}.

\subsection{Multi-Writer model(MW) for Persona Description Injection}

In the previous step, if the generated persona information is only provided as input prompts alongside story outlines, the LLM tends to focus more on the story context, neglecting the characters' detailed and diverse traits. To address this, we propose the \textbf{Multi-Writer model (MW)}, an additional module that takes persona information and previous draft as input to generate only specific descriptions about the persona. MW consists of two steps: (1) Multiple persona candidate generation and (2) Filtering and reranking.

\subsubsection{Multiple Persona Candidates Generation} During the draft step, each persona-writer generates a short description of different aspects of the protagonist, following the current paragraph. Each writer generates K candidates from the first-person perspective based on these instructions:  
\begin{enumerate}
\item (\textit{Relationship}) Describe thoughts/feelings about another person or the environment.
\item (\textit{Behavioral Habit}) Describe specific actions that reveal psychology. Usually, this candidate describes the protagonist's behavioral habit.
\item (\textit{Psychology}) Describe hint that reveals the protagonist's trauma. 
\item (\textit{Tone of Speech}) Describe dialogue that reveals the protagonist is speaking.
\item (\textit{Self-Description}) Describe appearance of the protagonist him/herself. 
\end{enumerate}

Full input prompts are available in Appendix \ref{sec:appendixa_5}. Specifically, in the Psychology candidate, incorporating the protagonist's trauma can add depth and richness to the character, thereby enhancing the overall narrative \cite{2018Tal}.

\subsubsection{Filtering and Ranking} Despite explicit instructions, persona-writers still focus more on previous draft than persona information, thereby failing to generate appropriate persona-descriptions. To filter out those bad samples, we calculate sentence-similarity between K samples and the persona information, selecting the sample with the highest similarity for each persona-writer. We also discard candidates that are too similar to previous outputs (average sum of self-BLEU 2 and 3 n-gram score > 0.0003). Remaining candidates are reranked using the Continuation-Score (\texttt{CS}).

\paragraph{Calculating Continuation-Score (\texttt{CS})} We fine-tune pretrained Qwen2-0.5B \cite{bai2023qwen} weights using an NVIDIA RTX A5000 for 2 hours on a regression task with ReedsyPrompts \cite{park2024longstory}. We select ReedsyPrompts due to its more suitable story length (average 2.4k) compared to other short story datasets such as WritingPrompts \cite{fan2018hierarchical} or ROCStories \cite{mostafazadeh2016corpus}. We use a 1k/0.1k/0.1k train/dev/test split.

First, we divide each story into paragraphs of a specific length. For golden label data, we concatenate two correctly ordered paragraphs; for negative label data, we concatenate two misordered paragraphs. Additionally, to create harder negative samples, we swap the first sentences of the posterior paragraphs in the negative samples with correct ones \citep{yang2022doc}. Qwen-0.5B learns to predict golden labels as 1 and negative labels as 0, quantifying how naturally two paragraphs connect. In the MW reranking process, we concatenate the current paragraph with each candidate, predict the \texttt{CS}, and select the candidate with the highest score. If no candidate exceeds the threshold (> 0.1), we discard all candidates. Finally, the best candidate is added to the current paragraph. As a result, we can insert detailed and diverse persona-descriptions into the story without compromising overall narrative quality by this process.

\subsection{Updating Persona Over Time}

The persona information can change as the story progresses and must be updated during drafting \citep{yang2022doc,he2024afspp,eger2017character}. At each completed node in the outline, we provide the current story and persona information to LLM, asking it to imagine how each trait has evolved. Example prompt can be found in Appendix \ref{sec:appendixa_6}.

\section{Experiments}

\subsection{Experimental Setup}

\paragraph{Task setup.}

Our task is to enhance the creativity of the story generation system. To this end, we utilize the IG to improve the diversity and concreteness of story elements, and MW to insert the detailed depiction of the protagonist into the story. We use both statistical analysis and human and LLM evaluation to evaluate whether creativity have increased by our methodology.
\paragraph{CCI Implementation Details.}
We implement CCI on the previous work, the DOC. We replace the Premise step of DOC with the IG module to generate story elements (see Appendix \ref{sec:appendixa_1},\ref{sec:appendixa_2} for more details). These story elements are then used in the same Plan step (Appendix \ref{sec:appendixa_4}) as in the original work. During the Draft step, we apply MW to inject persona descriptions at the end of each paragraph (Appendix \ref{sec:appendixa_5}). The similarity threshold and the CS score threshold  for the Multi-writer model were set empirically through repeated experiments. Persona updates are performed immediately after the completion of each outline node (Appendix \ref{sec:appendixa_6}). We use GPT-4o-2024-05-13 for all generations except during the Draft step. 
For the Draft stage, we conduct dual experiments using two types of LLMs: GPT-3.5-turbo-0125 and GPT-4o-2024-05-13.\\

\paragraph{Baselines.} 
  
Our primary baseline is DOC, because it is the only model that can generate stories of comparable length using the LLM pipeline without human intervention. However, we slightly modify the original version of DOC\footnote{We use the most recent code of DOC available at \url{https://github.com/facebookresearch/doc-storygen-v2}} to create a harder baseline. In the modified DOC, we upgrade the premise step of the original model using the `Text Only Story Elements Imagination' methods described in Appendix \ref{Text Only story elements Imagination}, further developing details through Section \ref{specification}, and then incorporating all of these story elements such as detailed persona, main plot, and background into the input context for the planning and drafting steps. Since this modification enhances the concreteness and diversity of the story elements, it can be considered an improved version of the original model. All settings, including the number of entities, outline nodes, and paragraphs per node, are shared with our framework.

Rolling is the simplest baseline, generating stories based solely on its last output\cite{yang2022re3}. It neither creates story elements nor outlines the entire story, starting directly from a short fixed prompt: ``Make a creative random story.''

\texttt{w/oIG} is identical to our framework but without the use of IG. In this version, we replace the IG step with text-only generation of story elements, while keeping all other components the same. Similarly, \texttt{w/oMW} does not apply MW, meaning that persona information generated during the IG step is incorporated only into the input context for the planning and drafting steps.

All baselines are implemented in two versions: \textit{-3.5} and \textit{-4o}. The \textit{-3.5} version uses GPT-3.5-turbo-0125 for drafting and the MW process, while the \textit{-4o} version utilizes GPT-4o-2024-05-13 for the same processes. 
\subsection{Main Experiment Results}
\label{Main_Experiment_Results}
\begin{table}[h!]
\centering
\setlength{\tabcolsep}{13pt} 
\renewcommand{\arraystretch}{1.0} 
\begin{tabular}{lcc}
\toprule
 & \textbf{ws(↓)} & \textbf{ss(↓)} \\ \midrule
\textbf{IG(DALL-E-3)} &\textbf{ 0.4723} & \textbf{0.474} \\
IG(SDM-v2) & 0.4996 & 0.5071 \\
IG(Illustrious-xl) & 0.5008 & 0.5995 \\
TO(GPT-4o) & 0.6607 & 0.6722 \\
\bottomrule
\end{tabular}
\caption{Comparison of word similarity (ws) and sentence similarity (ss) across different models. We use the same prompts for all baselines, except for text-only methods (TO). We bold the best values. See Appendix \ref{sec:appendix_Text2Img} for more details.}
\label{ws_ss}
\vspace{-5mm}
\end{table}

\begin{table}[h!]
\centering
\begin{tabular}{lccc}
\toprule
\textbf{Method} & \textbf{Emb-RV} & \textbf{LLM-RV} & \textbf{Sim(↓)}  \\ \midrule

\textbf{CCI-3.5} & \textbf{0.6581}  & \textbf{0.93}  & \textbf{0.7572} \\
DOC-3.5 & 0.6580 & 0.61  & 0.7651 \\ 
Rolling-3.5 & -  & -  & 0.7688 \\ \midrule

\textbf{CCI-4o} & \textbf{0.6696} & \textbf{0.88}  & \textbf{0.7586}\\
DOC-4o & 0.6553 & 0.68 & 0.7666 \\ 
Rolling-4o & -  & - & 0.7835\\
\bottomrule
\end{tabular}
\caption{Embbedding Relevance (Emb-RV), LLM Relevance (LLM-RV) and Similarity (Sim) for the final 20 outputs of CCI, DOC, and Rolling. The mean generation lengths for each model, across GPT-3.5 and GPT-4o, are as follows: CCI=2.1k, DOC=2.5k, Rolling=2.3k. We bold the best values.}
\label{Char-RV_Sim_Rep}
\vspace{-5mm}
\end{table}




\subsubsection{Diversity of Story Elements}
\label{Diversity of Story Elements.}
In this section, we analyze the repetitiveness of each story element (characters, background, and main plot) generated by various methods. We compare repetitiveness using two metrics: word similarity (WS) and sentence similarity (SS). To calculate WS, we average the 1-, 2-, and 3-gram BLEU \cite{papineni2002bleu} scores of all story elements. For SS, we average the cosine similarities of the sentence-BERT \cite{reimers2019sentence} embeddings for all story elements. Specifically, we take one story element (e.g., a character) from the entire set as the hypothesis, with the remaining samples (e.g., all other characters) as references. Then, we compute the word and sentence similarities for each reference and take the average. We repeat this process for each sample, using it as a hypothesis exactly once.

As shown in Table \ref{ws_ss}, all models with IG produce significantly lower repetition in both WS and SS compared to TO, demonstrating IG's ability to diversify story elements. We experiment with various state-of-the-art text-to-image models for visual imagination (\ref{Visual Imagination}), such as \texttt{stable-diffusion-2-1-base} and \texttt{Illustrious-xl-early-release-v0}, but DALL-E 3 turns out to be the most creative.


\subsubsection{Character-Relevance and Diversity of Full Stories}
To statistically compare the creativity of full stories, we present Character-Relevance and Similarity scores. Character-Relevance measures how well persona information is reflected in the story. Specifically, Embbedding-Relevance is calculated as the average cosine similarity between the embeddings of the persona information and the full story. For embedding long documents, we use OpenAI's large pretrained embedding model, \texttt{text-embedding-3-small}. Additionally, LLM-Relevance evaluates Character-Relevance through LLM, where GPT-4o rates the integration of persona information on a scale from 0 to 1. The example prompts are given in Appendix \ref{sec:appendix_LLM_Relevance}. The Similarity metric quantifies how similar each story is to others by calculating the cosine similarity for all pairs of full story embeddings and taking the average.

As shown in Table \ref{Char-RV_Sim_Rep}, CCI demonstrates better integration of persona information throughout the stories compared to DOC, as reflected in both Character-Relevance scores. The Similarity scores also suggest that our frameworks achieve greater diversity at a higher contextual level.


\subsubsection{Evaluating with Human and LLM Judgments}
\label{Evaluating with Human and LLM Judgments.}

\begin{table*}[ht]
\centering
\begin{tabular}{lccccc}
\toprule 
\texttt{} & \textbf{Char-Coherent} & \textbf{Char-Vivid} & \textbf{Concreteness} & \textbf{Novelty} & \textbf{Coherence} \\
\midrule
vs DOC-3.5  & 0.653 / 0.5    & 0.638 / \textbf{0.55}   & \textbf{0.833} / \textbf{0.6}    & 0.7916 / 0.5    & 0.75 / \textbf{0.6}    \\
vs \texttt{w/oIG}-3.5 & \textbf{0.833} / \textbf{0.575}  & 0.833 / \textbf{0.525}  & 0.75 / \textbf{0.625}    & 0.736 / \textbf{0.55}   & 0.889 / \textbf{0.575}  \\
vs \texttt{w/oMW}-3.5 & \textbf{0.875} / 0.5    & \textbf{0.833} / 0.45   & \textbf{0.944} / \textbf{0.625}  & \textbf{0.833} / \textbf{0.525}  & 0.722 / \textbf{0.6}    \\
\midrule 
vs DOC-4o   & \textbf{0.889} / \textbf{0.525}  & \textbf{0.889} / 0.5    & 0.778 / \textbf{0.7}  & 0.667 / \textbf{0.525}  & \textbf{0.944} / \textbf{0.65}  \\
vs \texttt{w/oIG}-4o  & 0.722 / \textbf{0.6}    & 0.611 / \textbf{0.475}  & 0.778 / \textbf{0.6}   & 0.555 / 0.5    & 0.611 / \textbf{0.525}    \\
vs \texttt{w/oMW}-4o  & \textbf{0.944} / \textbf{0.525}  & \textbf{0.944} / \textbf{0.575}  & \textbf{0.833} / 0.5  & 0.889 / 0.475  & 0.889 / 0.475    \\
\bottomrule
\end{tabular}
\caption{\textbf{Human / GPT-4o average win rates} across various criteria. For each score, the value to the left of the slash (/) represents the average preference ratio of human annotators, while the value to the right represents the preference ratio of the GPT-4o. The averages are calculated based on the annotators' choices for each of the 6 samples, where a score of 1 is given if our framework is preferred and 0 if the baseline is preferred. We bold the values where the win rate > 0.5 and the standard deviation < 0.2 for the human side, and where the win rate > 0.5 for the GPT-4o side.}
\label{main_exp_table}
\vspace{-3mm}
\end{table*}

\begin{table}[h!]
\centering
\begin{tabular}{lcc}
\toprule
\texttt{} & \textbf{Expression} & \textbf{Likability} \\ 
\midrule
vs \texttt{w/oMW}-4o & 0.683 / 0.5 & \textbf{0.6} / \textbf{0.575} \\
\bottomrule
\end{tabular}
\caption{\textbf{Human / GPT-4o average win rates} for Likability and Expression in the Multi-modal Interactive Story Generation experiments. We bold the values where the win rate > 0.5 and the standard deviation < 0.2 for the human side, and where the win rate > 0.5 for the GPT-4o side.}
\label{expression_likability}
\vspace{-3mm}
\end{table}

We conduct human evaluations with eight highly educated participants who have a strong interest in literature (see Appendix \ref{sec:appendixF} for more details of human annotators). In each experiment, annotators compare our framework with a baseline model by observing 6 full samples (including background, mainplot, character, and the full story) from each, and for each sample, they separately determine which performs better based on the following metrics.\footnote{You can see all the samples and rubrics used for evaluation at \url{https://www.2024cci.p-e.kr/human_experiments.html}} The evaluation metrics are as follows:

\begin{enumerate} \item \textit{Character-Coherent.} Which model's final output better maintains the character's personality? \item \textit{Character-Vivid.} Which model's final output depicts the character more vividly? \item \textit{Concreteness.} Which model's story elements are more concrete? \item \textit{Novelty.} Which model's story elements are more novel? \item \textit{Coherence.} Which model's final output has a more coherent story? \end{enumerate}

For LLM evaluation, we conduct the same experiment with the LLM (GPT-4o), using each model's 20 stories. To eliminate location bias, we run two experiments for each sample, swapping the order of the two stories in the prompt, and then average the preference ratios from both experiments. The overall human and LLM evaluation results are presented in Table \ref{main_exp_table}, with the corresponding average annotator agreements provided in Appendix \ref{sec:appendix_agreement}.

\paragraph{vs DOC.} The average total win rates are 0.794 ± 0.193 for CCI-3.5 and 0.796 ± 0.22 for CCI-4o. Annotators' preference for CCI-3.5 is stronger and more  confident than the preference for CCI-4o, likely because GPT-4o's advanced generation capabilities partially cover the benefits of the IG and MW modules. Nevertheless, CCI outperforms DOC across nearly all metrics for both human and LLM evaluations.

In particular, CCI-3.5 excels in Concreteness, while CCI-4o shows significant improvements in Character Coherence and overall Coherence in both human and LLM evaluations, even within the human evaluation threshold of std < 0.2. This demonstrates that CCI not only enhances the consistency and concreteness of character identities and other story elements, but also improves overall story coherence, even though improving coherence is not an explicit goal of our work.

\paragraph{vs \texttt{w/oIG}, \texttt{w/oMW}.}
As the absence of IG negatively impacts Concreteness and Novelty, CCI is preferred in terms of these metrics over \texttt{w/oIG}. This confirms the IG module's effectiveness in diversifying story elements. Furthermore, IG enhances the quality of the final stories by allowing the incorporation of high-quality story elements, particularly for characters. 

The results of \texttt{w/oMW} reveal MW's crucial role in enriching persona descriptions and improving overall story quality. Although \texttt{w/oMW} also utilizes IG, CCI is still preferred in terms of Concreteness and Novelty in human evaluations, as annotators assess not only `how concrete or novel each story element is,' but also `how effectively these elements are utilized in the final story in terms of concreteness and novelty.' The MW module's active injection of persona descriptions enhances the results from this perspective.

\subsection{Multi-modal Interactive Story Generation}

In this experiment, the annotators provide their own images for the protagonist, background, and main plot, then respond with their preference between the final output of our framework and the \texttt{w/oMW}. The GPT-4o also participates in the evaluation, given the same stories and matched custom images. The evaluation metrics are as follows:
\begin{enumerate}
\item \textit{Expression.} Which model's final output better expresses the provided images?
\item \textit{Likability.} Which model's final output is better?
\end{enumerate}

Appendix \ref{appendix:Multimodal Interactive Story Generation} shows an example of a multimodal interactive custom story. We use custom images provided by an annotator and transform the images into text descriptions through the Image-Guided Story Elements Creation (\ref{Image-guided story elements creation}) and use them in the rest of the steps. Specifically, the features are interwoven and extended together through the Specification process (\ref{specification}), and MW utilizes them to insert multiple persona descriptions into the story. Since the \texttt{w/oMW} does not directly inject persona information, human annotators answer that our framework better expresses the image, and all evaluators, including both humans and GPT-4o, generally prefer our framework, as shown in Table \ref{expression_likability}.  This demonstrates that MW enhances the incorporation of user-provided images into the story, suggesting that future research in multimodal interactive story generation should consider utilizing the MW module to improve quality.

\section{Conclusion}

We propose CCI, a framework that enhances creativity in story generation by using IG to generate visual representations of key story elements and MW to insert detailed, context-appropriate descriptions of protagonists. CCI improves story diversity and character depth without requiring extensive training, such as fine-tuning LLMs. Both human and LLM evaluations favor CCI over baselines. Additionally, CCI can be applied to multi-modal interactive story generation, demonstrating improvements in both quantitative and qualitative aspects.

\section{Limitations}

Our two main contributions, IG and MW, are model-agnostic. However, we cannot test whether they improve other models due to the lack of other LLM-based recursive long story generation systems. In the future, if other LLM-based story generation frameworks are proposed, it will be possible to verify whether our main contributions are truly model-agnostic or not.

We arbitrarily selected the personal traits to be incorporated into the stories. Although traits such as personality, habits, speech tone, and appearance have been widely used in past works, they are chosen by our intuition. In reality, the number of personal traits is countless, and it is unclear which traits are optimal for use in a story. Future works need to establish more precise criteria for selecting personal traits or include a greater variety of personal traits.

\section{Ethical considerations}

Since our work focuses on adding diverse and detailed imagination to fictional stories, we are almost free from the risk of generating untruthful text. However, we need to consider the potential risk of developing toxic content. Many great literary works include sexual or violent depictions to express the darker aspects of humanity. Consequently, our efforts to enhance the imaginative details of stories might lead to including violent or sexual descriptions. Fortunately, our work uses OpenAI's ChatGPT and DALL-E 3 for language and image generation, which strictly censor violent or sexual content. However, if other models are used, they might bypass such censorship, requiring future researchers to exercise moral judgment.

\bibliography{custom}

\clearpage

\appendix

\section{Survey of Weaknesses of AI's stories}
\label{sec:survey}

Before conducting the main experiments, we asked all annotators who participated in the human evaluations in Section \ref{Evaluating with Human and LLM Judgments.} to choose the most important factors that distinguish AI-generated stories from human-written novels. The choices are as follows:

\begin{enumerate}[label=A\arabic*.]
\item Lack of coherence compared to human level.
\item Lack of detailed description compared to human level.
\item Lack of inner human depiction compared to human level.
\item Lack of sentence construction compared to human level.
\item Lack of engaging plot compared to human level.
\item Lack of originality compared to human level.
\end{enumerate}

Participants actively engaged in our research by selecting up to 3 choices and providing other comments freely. As shown in Table \ref{table:votesdist}, they perceive the greatest differences to be the lack of inner human depiction and detailed description compared to human-level writing, while also finding it less interesting and showing low originality.
\begin{table}[h!]
\centering
\begin{tabular}{|c|c|c|c|c|c|c|}
\hline
 & A1 & A2 & A3 & A4 & A5 & A6 \\ \hline
votes & 0 & 4 & 5 & 0 & 1 & 2 \\ \hline
\end{tabular}
\caption{Votes Distribution. The A3 receives the most votes, followed by A2. Additional comments from participants include: \textit{1. Lack of dimensionality in character personality. 2. Repetition in descriptions.}}
\label{table:votesdist}
\end{table}
\section{Term of uses}
This research utilized OpenAI's GPT-3.5-turbo-0125, GPT-4o-2024-05-13 and DALL-E 3 for academic purposes. The usage of these models complied with OpenAI's [Terms of Use](\url{https://openai.com/terms}) and [Usage Policies](\url{https://openai.com/policies/usage-policies}).

\section{Details of Human Evaluation}
\label{sec:appendixF}
We recruited graduate and undergraduate students fluent in English through the university’s literature club. The recruited annotators were provided with a detailed description of task definitions, instructions, and samples of each model. Also, all applicants were informed that their annotations would be used for academic purposes and would be published in paper material through the recruitment announcement and instructions. \\ A total of 8 annotators were provided with 6 full samples including title, character, background, and main plot from each model and responded with their preferred model. You can find the samples used for the survey at \url{https://www.2024cci.p-e.kr/}. For the payment of the annotators, the co-authors conducted annotations for 5 hours first to estimate the average number of annotations that could be completed in the same time. Based on this estimation, a rate of 1.83 dollars per example was established to ensure that the annotators would be paid at least the minimum wage. 

\section{Detailed Hyperparameters and Conditions} \label{sec:appendixf} For GPT-3.5-turbo, we set the hyperparameters as frequency penalty=1, presence penalty=0, top p=0.99, and temperature=1. To avoid the increase in perplexity due to high temperature, we did not increase it. The hyperparameters for GPT-4o-2024-05-13 are the same as for GPT-3.5-turbo-0125. For the plan stage, the hyperparameters are set as follows: outline's max depth=2, min children=2, preferred max children=4, max children=5. In the draft stage, the important hyperparameters are 'min passages per node=1, max passages per node=2' for CCI, and 'min passages per node=1, max passages per node=3' for DOC and \texttt{w/oMW}. Additionally, since the Detailed Controller from the original work\cite{yang2022doc} requires OPT-175B, we do not use the controller due to a lack of GPU memory. These conditions are shared across all the models in this paper.

\section{Annotator's Agreement}
\label{sec:appendix_agreement}
Given the inherent asymmetry in our data, where our model is superior and consistently favored by respondents, traditional metrics like Cohen’s Kappa may underestimate agreement due to the high expected agreement. To address this, we employed Gwet’s AC1, a more robust measure less sensitive to data imbalance, ensuring that the observed trends accurately reflect the statistical reliability of the results.

As shown in Table \ref{tab:gwets_ac1}, the Gwet's AC1 values indicate at least moderate agreement (>0.4), demonstrating sufficient inter-rater consistency and confirming the reliability of our evaluation results.

\begin{table}[ht]
\centering
\begin{tabular}{lc}
\toprule
\textbf{Comparison}                & \textbf{Average} \\ \midrule
CCI-3.5 vs DOC                     & 0.6070           \\
CCI-3.5 vs wo/IG                   & 0.6762           \\
CCI-3.5 vs wo/MW                   & 0.4991           \\
CCI-4o vs DOC                      & 0.5298           \\
CCI-4o vs wo/IG                    & 0.6056           \\
CCI-4o vs wo/MW                    & 0.5064           \\
\midrule
(Exp / Lik) CCI-4o vs wo/MW   & 0.5000           \\ \bottomrule
\end{tabular}
\caption{Gwet's AC1 Analysis Averages.}
\label{tab:gwets_ac1}
\end{table}

\section{Example Prompts}
\label{sec:appendixa}
\subsection{Image-guided story elements Imagination}
We use DALL-E 3 to receive an image for image-guided story elements imagination. Then, we input the generated image into GPT-4o-2024-05-13 to create a description of a character, background, or climax. We observe that allowing too fantastic or far from reality imagination led to the repetition of similar themes (e.g., cliffs, dragons, magic), and thus restrict it. Additionally, we conduct experiments from Appendix \ref{sec:appendixE} to find the optimal prompt for maximizing creativity in DALL-E 3.

\label{sec:appendixa_1}
\onecolumn
\begin{table}[h!]
\centering
\begin{tabular}{|p{15cm}|}
\hline
\textbf{Image-guided Character Imagination} \\ \hline
1. DALL-E 3 prompt: A character from random genre of manga. \\

2. GPT-4o-2024-05-13 prompt : \\Look carefully this image, and give me your imagination of detailed description of appearance of the character, and his or her name. in 8 sentences or less of it. Feel free to use your imagination to the fullest. Don't make any other explanations, questions, or comments. Answer format :  (Don't mimic this example) : Alex : a lean, pale boy. he limps. Must keep the format.
\\ \hline
\textbf{Image-guided Background Imagination} \\ \hline
1. DALL-E 3 prompt :  A background from random genre of movie. \\

2. GPT-4o-2024-05-13 prompt :  \\Look carefully this image, and give me detailed description of it in 8 sentences or less of it. Feel free to use your imagination to the fullest. Don't make any other explanations, questions, or comments.
\\ \hline
\textbf{Image-guided Climax Imagination} \\ \hline
1. DALL-E 3 prompt : A climax of random genre of movie. \\

2. GPT-4o-2024-05-13 prompt : \\Look carefully this image, and give me detailed description of what is happening in 8 sentences or less of it. Feel free to use your imagination to the fullest. Don't make any other explanations, questions, or comments.
\\ \hline
\end{tabular}
\caption{These are prompts to generate a character, background, and climax guided by an image. We provide prompts for DALL-E 3 to generate images and for GPT-4o-2024-05-13 to create text descriptions based on those generated images.}
\end{table}
\twocolumn
\subsection{Specification Persona and Main plot}
\label{sec:appendixa_2}
In the specification process, we elaborate on two story elements: the persona and the main plot. The questions for detailing the persona include: 1. Dark secret. 2. Family environment. 3. Description his or her appearance oneself. 4. Specific way of speakings(tone of speech). 5. Personality.\citep{goldberg2013alternative} 6. The most significant events in life. 7. Habitual behaviors. We selected this list based on our intuition. The previously generated background is also used as a prompt in the specification.

For the main plot specification, we follow a structured process. We perform the entire process of the Chain of Asking why. This process involves asking 'why' repeatedly, up to the maximum value M, or repeating it until the response to STEP 3 is 'No'. We then augment 'why happen' with the evidence found in STEP 2, creating a logical and coherent plot.
\onecolumn
\begin{table}[h!]
\centering
\begin{tabular}{|p{15cm}|}
\hline
\textbf{Protagonist's personal traits specification} \\ \hline
[\textbf{name}] is your name, and you are [\textbf{description}]. \\
The main background where you live in : [\textbf{background}] \\
Now, [\textbf{name}], with your own self-portrait, and use your utmost imagination to answer the question.
Don't make any other suggestions, questions, or comments. \\
Question. \\

\begin{enumerate}
\item \textbf{\textit{What is your dark secret? Don't imagine too far from reality or fantastic, Do imagine realistically and concretely.
\item What is your family environment like? Don't imagine too far from reality or fantastic, Do imagine realistically and concretely.
\item Describe your appearance in detail. Don't imagine too far from reality or fantastic, Do imagine realistically and concretely.
\item What is your specific way of speakings(tone of speech)? Answer mainly with examples where the characteristic is actually applied.
\item What is your personality like? Personality is one of the following five traits: extraversion (EXT), agreeableness (AGR), conscientiousness (CON), neuroticism (NEU), or openness to experience (OPE). Answer mainly with examples where the personality is actually applied. Don't imagine too far from reality or fantastic, Do imagine realistically and concretely.
\item What are the most significant events you have experienced from childhood to now, primarily those that left a **trauma**? Do not write the number before the answer. Don't imagine too far from reality or fantastic, Do imagine realistically and concretely.
\item What is very specific habitual behaviors? list 5. Do imagine realistically and concretely.
}}

\end{enumerate}

\\ \hline
\end{tabular}
\caption{These are prompt to specify the protagonist's personal traits. We use story elements generated in \ref{sec:appendixa_1} to fill in [\textbf{name}], [\textbf{description}], [\textbf{background}], and \textit{other person}. }
\end{table}

\begin{table}[h!]
\centering
\begin{tabular}{|p{15cm}|}
\hline
\textbf{Chain of Asking Why} \\ \hline
STEP 1. Find out missing backgrounds.\\\\
Text : [\textbf{main plot}]
\\\\Find out the most important missing backgrounds or naturally arising questions in the given text, at most 3.
If there are ambiguous terms like `secret', `hidden ...', `something', `someone' or `conspiracy' that do not reveal detailed information, you must ask what their true nature is.\\\\
STEP 2. Imagine evidences of missing backgrounds.\\\\
You are [\textbf{name}]. A event recently taking place is [\textbf{main plot}].\\Below is the statements you provided about yourself : [\textbf{personal-traits}]\\
Now, use your imagination to the fullest, imagine evidences of [\textbf{missing backgrounds}]\\\\
STEP 3. Check if any missing backgrounds remains.\\\\
Text : [\textbf{main plot}] [\textbf{Evidences}]\\\\
Are there ambiguous terms like `secret', `hidden ...', `something', `someone', `unresolved' or `conspiracy' that do not reveal detailed information in the given text? If there so, please answer `Yes.' If there aren't, please answer `No'.

\\ \hline
\textbf{Main plot Specification} \\ \hline
You are [\textbf{name}]. A event recently taking place is [\textbf{main plot}].\\Below is the statements you provided about yourself : [\textbf{personal-traits}] \\
Based on statements about yourself, answer this question. Don't imagine too fantastical or far from reality.
\begin{enumerate}
\item Now, this main event is an inevitable destiny for you. Based on your statements, imagine why such an inevitable event of fate has befallen you. don't imagine too fantastical or far from reality.
\item How do you react to this event? MUST Based on statements about yourself, answer this question as convincingly as possible.
\item Summarize the whole plot in 5 sentences, which starts from `the story of ...' 
\end{enumerate}
\\ \hline

\end{tabular}
\caption{These are prompts to specify the main plot. We use main plot and personal traits generated in \ref{sec:appendixa_1} and \ref{sec:appendixa_2} to fill in [\textbf{main plot}], [\textbf{personal-traits}].  }
\end{table}
\twocolumn
\subsection{Planning and Drafting with story elements}
\label{sec:appendixa_4}
Story elements are used as prompt elements in the plan and draft stages. This prompt utilizes the implementation from the previous work(\url{https://github.com/facebookresearch/doc-storygen-v2}).
\onecolumn
\begin{table}[ht]
\centering
\begin{tabular}{|p{15cm}|}
\hline
\textbf{Prompts for plan} \\ \hline
Premise: [\textbf{main plot}] Setting: [\textbf{background}] Characters and Entities: [\textbf{characters}] Write a very brief, high-level outline for this story...\\Remember, You are [\textbf{name}]. Must write in the first-person perspective (using ``I'').

\\ \hline
\textbf{Prompts for draft} \\ \hline
I will give you the high-level premise for a book, along with a summary of what has happened so far. This book is being authored by a well-known novelist(you), who received glowing reviews from critics, with praise for the interesting dialogue and interactions between characters. Premise: [\textbf{main plot}] Relevant context: [\textbf{desc}][\textbf{personal-traits}]...\\
Remember, You are [\textbf{name}]. Must write in the first-person perspective (using ``I'').

\\ \hline
\end{tabular}
\caption{These are prompts to plan and draft. We use specified story elements generated in \ref{sec:appendixa_1} and \ref{sec:appendixa_2} to fill in [\textbf{name}],[\textbf{desc}],[\textbf{background}],[\textbf{main plot}] and [\textbf{personal-traits}]. Unchanged prompts from the existing implementation are omitted.}
\end{table}
\twocolumn
\subsection{Persona Injection by Multi-writer model}
\label{sec:appendixa_5}
Persona insertion by the Multi-writer model includes five types of character descriptions. 1. Thoughts or feelings about others (Relationship). 2. Behavior reveals psychology(Behavioral Habit), 3. Hint that reveals trauma (Psychology). 4. Dialogue (Tone-of-speech). 5. Description of appearance(Self-description). Each answer made separately, with 8 candidates for each question. These candidates are then used to filter and rerank persona candidates. Ultimately, the candidate with a continuous score above a certain threshold and the highest rank is inserted into the draft.
\onecolumn
\begin{table}[h!]
\centering
\begin{tabular}{|p{15cm}|}
\hline
\textbf{Multiple Persona Candidates Generation.} \\ \hline
You are [\textbf{name}]. Your characteristics : [\textbf{personal-traits}]. Deeply Considering your characteristics and the current context, 
\begin{enumerate}
\item  \textit{\textbf{Describe your thoughts/feelings about another person or the environment.}}
\item \textit{\textbf{Describe specific action that reveals yours psychology.}}
\item \textit{\textbf{Describe hint that reveals yours trauma.}} 
\item \textit{\textbf{Describe **dialogue** that reveals you're speaking}}.
\item \textit{\textbf{Describe appearance of yourself}}. 
\end{enumerate}
Must write in the first-person([\textbf{name}]) perspective (using I). [\textit{Current Context}]
\\ \hline
\end{tabular}
\caption{This is an example of multiple persona candidate generation. In this example, we use story elements and personal traits generated in previous steps. 8 candidates are generated for each question, and these candidates are then used to filter and rerank persona candidates.}
\end{table}
\twocolumn

\subsection{Updating Persona Over Time}
\label{sec:appendixa_6}
Each time an event node is completed into a story, we update the persona information, which are then used in the next persona injection. To update the persona, we use the story of the currently written outline node and the current personal traits as the current context to modify the personal traits. To avoid overly repetitive descriptions throughout the story, we ensure that some traits are prompted to change each time.

\subsection{Instruction for LLM Relevance score}
\label{sec:appendix_LLM_Relevance}
An example prompt to ask LLM to calculate the relevance score is in Table \ref{table:LLM_Relevance}.
\onecolumn
\begin{table}[h!]
\centering
\begin{tabular}{|p{15cm}|}
\hline
\textbf{Updating Persona Over Time.} \\ \hline
You are [\textbf{name}]. Current situation : [\textit{Current Context}] \\
Based on all this information, answer the question.
\begin{enumerate}
\item  \textit{\textbf{How has your specific appearance changed or newly developed due to the given context, given that it has changed?}}
\item \textit{\textbf{How has your specific way of speaking (specific favorite words, speech habits, etc.) changed or newly developed due to the given context, given that it has changed?}}
\item \textit{\textbf{How has your personality changed or newly developed due to the given context, given that it has changed?}}
\item \textit{\textbf{Have you developed any specific trauma due to the given context? If you have, describe the future event that might trigger it and how it might manifest, and if you have not, just respond with the previous answer.}}
\item \textit{\textbf{How have your specific habitual behaviors (e.g., biting your nails, running your fingers through your hair, etc.) changed or newly developed given the context, given that they have changed?}}
\item \textbf{\textit{How have your feelings or thoughts about other person changed or newly developed?}}

\end{enumerate}
Don't make any other suggestions, questions, or comments.
\\ \hline
\end{tabular}
\caption{This is an example prompt for persona updating. Each time an event node is completed into a story, we update the personal traits, which are then used in the next Persona injection.}
\end{table}

\begin{table}[h!]
\centering
\begin{tabular}{|p{15cm}|}
\hline
\textbf{Instruction for LLM Relevance scoring.} \\ \hline
Evaluate how well the following trait of the character is reflected in the given story on a scale of 0 to 1, where 0 means the trait is not reflected at all, and 1 means the trait is perfectly reflected. Trait: [\textit{trait}] Story: [\textit{story}].
\\ \hline
\end{tabular}
\caption{This is an example prompt for LLM Relevance scoring.}
\label{table:LLM_Relevance}
\end{table}
\twocolumn

\section{Text Only story elements Imagination}
\label{Text Only story elements Imagination}
This is an example prompt for generating story elements using text only, without IG. We find that allowing overly fantastical imagination leads to the repetition of similar themes (e.g., cliffs, dragons, magic) and thus restricts it. Additionally, when generating the main plot, we made the prompt imagine a single dramatic event. To broaden the scope of imagination, we permit using adult elements. Finally, we include a sentence encouraging maximum use of imagination in all creations (e.g., Feel free to use your imagination to the fullest.)

\onecolumn
\begin{table}[h!]
\centering
\begin{tabular}{|p{15cm}|}
\hline
\textbf{Text Only Character Imagination} \\ \hline
Make name and appearance of a main character.
Don't answer with something too fantastical or too far from reality. Don't imagine attributes beyond the appearance. Feel free to use your imagination to the fullest. Answer Format(Don't copy the same.) : ``Alex : a lean, pale boy. he limps. '' Must keep the format.
\\ \hline
\textbf{Text Only Background Imagination} \\ \hline
Descripe ONE main background of a story. Feel free to use your imagination to the fullest. Don't answer with something too fantastical or too far from reality.
\\ \hline
\textbf{Text Only Climax Imagination} \\ \hline
Imagine the most dramatic event of a story. Use your imagination to the fullest. This dramatic event may contain elements of an adult production. Don't answer with something too fantastical or too far from reality.
\\ \hline
\end{tabular}
\caption{These are prompts to generate a character, background, and climax without an image. }
\end{table}
\twocolumn

\section{Best text2img model for IG}
\label{sec:appendix_Text2Img}
We experiment with three types of text-to-image models for IG: DALL-E 3, stabilityai/stable-diffusion-2-1-base, and OnomaAIResearch/Illustrious-xl-early-release-v0. The other baseline is TO, which is the result of text-only story element imagination. To calculate `ws', we average the sum of 2- and 3-gram BLEU scores of story element pairs. For `ss', we average the cosine similarities of Sentence-BERT embeddings of story element pairs. We report the average values of Char-ws, Back-ws, and Climax-ws as `ws', and Char-ss, Back-ss, and Climax-ss as `ss' in Table \ref{ws_ss}. As shown in Table \ref{appendix_ws_ss}, DALL-E 3 demonstrates the most powerful imaginative ability.

\section{DALL-E 3's Best Prompt}
\label{sec:appendixE}
We experiment with three types of prompts to maximize the creativity of DALL-E 3.
\begin{enumerate} [label=Type \arabic*., leftmargin=3em]
    \item ``a character/background/climax from a random movie.'' 
    \item ``a character/background/climax from a random story.'' 
    \item ``a character/background/climax from a random manga.''
\end{enumerate} 
As shown in Table \ref{table:wholeDALL-E}, the overall diversity of Characters is the greatest in Type 3, and the Main plot and Background are the best in Type 1. Therefore, the results in Table \label{sec:appendix_Text2Img} are the average BLEU 1, 2, and 3 n-grams and the SBERT scores for Characters made from Type 3 and Main plots and Backgrounds made from Type 1 prompts.
\onecolumn

\begin{table*}[h]
\centering
\begin{tabular}{lcccccc}
\hline
 & \textbf{Char-ws} & \textbf{Char-ss} & \textbf{Back-ws} & \textbf{Back-ss} & \textbf{Climax-ws} & \textbf{Climax-ss} \\ \hline
\textbf{IG(DALL-E-3)} & 0.4679 & \textbf{0.3993} & \textbf{0.4643} & 0.5915 & 0.4847 & \textbf{0.4313} \\
IG(SDM-2) & 0.5318 & 0.4630 & 0.4865 & \textbf{0.5540} & 0.4808 & 0.5045 \\
IG(Illustriouss-xl) & \textbf{0.4272} & 0.4403 & 0.6119 & 0.7807 & \textbf{0.4633} & 0.5655 \\
TO & 0.7400 & 0.6669 & 0.6271 & 0.6695 & 0.6152 & 0.6803 \\
\hline
\end{tabular}
\caption{The repetitiveness of story elements created by various text2img modules and Text-only method.}
\label{appendix_ws_ss}
\end{table*}

\begin{table}[h!]
\centering
\renewcommand{\arraystretch}{1.1} 
\begin{tabular}{>{\raggedright\arraybackslash}m{4cm} >{\centering\arraybackslash}m{2cm} >{\centering\arraybackslash}m{2cm} >{\centering\arraybackslash}m{2cm} >{\centering\arraybackslash}m{2cm}}
\hline
\rowcolor[HTML]{D3D3D3} 
\multicolumn{5}{c}{\textcolor{blue}{\textbf{Character}}} \\ \hline
\textbf{} & \textbf{sb1} & \textbf{sb2} & \textbf{sb3} & \textbf{ssbert} \\ \hline
Text only & 0.7886 & 0.6036 & 0.4556 & 0.6435 \\ 
type1 & 0.7463 & 0.5382 & 0.3448 & 0.3825 \\ 
type2 & 0.7023 & 0.4707 & 0.2668 & \textbf{0.3782} \\ 
\textbf{type3} & \textbf{0.6951} & \textbf{0.4496} & \textbf{0.259} & 0.3993 \\ \hline

\rowcolor[HTML]{D3D3D3} 
\multicolumn{5}{c}{\textcolor{blue}{\textbf{Background}}} \\ \hline
\textbf{} & \textbf{sb1} & \textbf{sb2} & \textbf{sb3} & \textbf{ssbert} \\ \hline
Text only & 0.7667 & 0.5057 & 0.3238 & 0.6819 \\ 
\textbf{type1} & \textbf{0.7183} & \textbf{0.4437} & \textbf{0.231} & \textbf{0.5915} \\ 
type2 & 0.7543 & 0.5085 & 0.3013 & 0.6725 \\ 
type3 & 0.7721 & 0.5184 & 0.3078 & 0.6528 \\ \hline

\rowcolor[HTML]{D3D3D3} 
\multicolumn{5}{c}{\textcolor{blue}{\textbf{Climax}}} \\ \hline
\textbf{} & \textbf{sb1} & \textbf{sb2} & \textbf{sb3} & \textbf{ssbert} \\ \hline
Text only & 0.7489 & 0.4891 & 0.29845 & 0.6602 \\ 
\textbf{type1} & \textbf{0.7219} & \textbf{0.4651} & \textbf{0.267} & \textbf{0.4313} \\ 
type2 & 0.7315 & 0.4726 & 0.2842 & 0.4616 \\ 
type3 & 0.7542 & 0.513 & 0.3109 & 0.5634 \\ \bottomrule
\end{tabular}


\caption{The average self-BLEU 1, 2, and 3 n-grams, and self-SentenceBERT scores for 20 Character, Background, and Main plot samples generated from Text-only method, and Type 1, Type 2, and Type 3 DALL-E-3 prompts. For each column, we bold the lowest repetitiveness and thus highest diversity score.}
\label{table:wholeDALL-E}
\end{table}
\twocolumn

\section{Example of DOC}
\label{sec:appendixe}
We report the three final results of DOC-4o that is used in human and LLM evaluation. In this baseline, story elements are generated using text-only methods instead of IG. The Specification stage is conducted in the same manner as CCI. However, the detailed story elements are used solely as prompts, not being used by DOC. 

In the case study, we find the Text-only imagined story elements generally reproduce the very similar themes. Furthermore, the descriptions of characters are often limited to age, hair, and clothes, indicating a lack of imagination. In the stories, while some sentences do describe the inner thoughts or vivid actions of characters, these descriptions are made to advance the context, and personal traits are hardly utilized. As a result, readers obtain few clues about the distinctive characteristics of the protagonist, which make it hard to imagine the character vividly.

\onecolumn
\begin{table}[h!]
\centering
\begin{tabular}{|p{15cm}|}
\hline
\textbf{Text-only imagined character example 1} \\ \hline
Elena : 27 years old. a slender, ghostly woman. I wear a long, tattered dress. My hair is dark and tangled.
\\ \hline
\textbf{Text-only imagined background example 1} \\ \hline
Dramatic Background: The setting is a crumbling castle perched atop a fog-covered cliff, brooding under a perpetually overcast sky. Lightning frequently flashes, casting eerie shadows across ancient stone walls lined with faded tapestries. The sound of waves crashing below echoes throughout the empty corridors, enhancing the sense of desolation. Vines creep through cracked windows, reclaiming the neglected structure where secrets are whispered by the wind. Only distant chimes from an old bell tower pierce the haunting silence envelop
\\ \hline
\textbf{Text-only imagined main plot example 1} \\ \hline
Amidst a violent thunderstorm, the castle's walls tremble as portraits crash to the ground. A lone figure stands weeping at the edge of a crumbling parapet, rain drenching her tattered gown. A long-lost lover emerges from shadow, his ghostly form illuminated by lightning. Their anguished voices echo through empty halls while the wind howls their broken vows into eternity. As they reach out, a lightning strike shatters stone, sending one spiraling into oblivion below. An anguished cry reverberates through desolate corridors as darkness consumes what remains of shattered dreams and love lost forever.
\\ \hline
\end{tabular}
\caption{This is the first example of Text-Only story element generation. The characters are less concrete than IG's results.}
\end{table}

\begin{table}[h!]
\centering
\begin{tabular}{|p{15cm}|}
\hline
\textbf{DOC partial example 1} \\ \hline
Hi, I am Elena. The storm rages on, pounding the castle like relentless waves against jagged cliffs. My breath catches in my throat as I choke back sobs that mingle with the howling wind. The parapet feels ice-cold beneath my feet as raindrops carve rivers down my pallid cheeks The storm rages on, pounding the castle like relentless waves against jagged cliffs. My breath catches in my throat as I choke back sobs that mingle with the howling wind. The parapet feels ice-cold beneath my feet as raindrops carve rivers down my pallid cheeks. Lightning illuminates the sky , casting eerie shadows that dance upon the ancient stones. I cling to the cold, rough edge of the parapet, my fingers trembling and white-knuckled. The thunder roars above, as if nature itself mourns with me. There he stands—a spectral figure cloaked in mist and sorrow. His eyes mirror my suffering, their once vibrant hue now clouded with the sadness of years long past. The wind whips through his translucent form, blending rain and tears as one. My heart aches with the unbearable weight of both reunion and loss. ``Why?'' I cry out, my voice barely more than a whisper carried away by the storm. ``Why did you leave me in this abyss of sorrow?'' His spectral eyes, once brimming with love, now hold only the faintest glimmer of recognition. ``Elena,'' he murmurs, his voice an ethereal echo of what it once was, ``I never intended to abandon you.'' ``I never intended to abandon you.'' My heart wrenches as I listen to his words, barely audible over the howling wind. ``But you did,'' I whisper, my voice breaking like the waves crashing below. ``You left me here in this fortress of desolation, surrounded by shadows and specters of what once was.'' His form wavers like a candle flame, and for a moment, I fear he might fade completely. The cold wind screams through the parapet, bringing with it the scent of salt and decay. `It was not by choice,'' he replies, his voice trembling as though grappling with a reality that once was. His form wavers like a candle flame, and for a moment, I fear he might fade completely. The cold wind screams through the parapet, bringing with it the scent of salt and decay. `It was not by choice,'' he replies, his voice trembling as though grappling with a reality that refuses to be forgotten. My hand reaches out, trembling, to touch his cold visage, but it passes through air tinged with despair. `Why did you leave me?'' I whisper, my voice barely audible over the storm's rage. Thunder drowns his reply; yet in the flash of lightning that follows I see the anguish in his eyes. His spectral hand reaches back, mirroring my own futile gesture. The storm's relentless fury seems to amplify our shared agony, the wind pulling at my hair and dress as if trying to separate us further. `Elena,'' he murmurs, a sound barely more than the breath of the storm itself, yet it slices through me like a blade. `Why did you leave me?'' I cry out, my voice mingling with the howling wind. The rain burns on my skin like tears of the heavens themselves, each drop a reminder of my unending sorrow. He moves closer, his ethereal form shimmering against the stark light of the lightning. `I never left you,'' he murmurs, his voice a hollow echo that drifts through the desolate library like ghostly fingers brushing over old tomes. Our eyes meet, and for a moment, time ceases in its relentless march. His gaze holds me captive, and I am lost in the depths of his sorrow as if peering into a mirror reflecting my own despair. His touch is cold, a phantom sensation that sends shivers down my spine yet offers the faintest trace of comfort amidst the torment. ``Every shadow in this castle carries your scent, every whisper me captive, and I am lost in the depths of his sorrow as if peering into a mirror reflecting my own despair. His touch is cold, a phantom sensation that sends shivers down my spine yet offers the faintest trace of comfort amidst the torment.  (...)
\\ \hline
\end{tabular}
\caption{This is part of the first final result of the DOC. The full generated result (about 2.5k) is too long, so we only report a portion of the full sample. The story is coherently formed, but there is little evidence of who the protagonists themselves are, and the persona of the character in the prompt is barely revealed. You can see the full story in \url{https://www.2024cci.p-e.kr/4o/vsDOC/view.html?n=3}}
\end{table}

\begin{table}[h!]
\centering
\begin{tabular}{|p{15cm}|}
\hline
\textbf{Text-only imagined character example 2} \\ \hline
Elena: 25 years old. a tall woman with raven-black hair. I wear a tattered cloak.

\\ \hline
\textbf{Text-only imagined background example 2} \\ \hline
 A towering castle loomed atop a misty hill, shrouded in swirling fog. Cracks ran along its weathered stone walls, whispering tales of old battles. Dark clouds perpetually gathered above, casting eternal shadows over the ancient grounds. A dense forest enveloped the castle, hiding secrets in its twisted branches and haunting echoes. Nearby villagers spoke in hushed tones about eerie lights flickering at windows long abandoned to time.
\\ \hline
\textbf{Text-only imagined main plot example 2} \\ \hline
In the castle's grand hall, Lady Elara faced her brother, Lord Cedric, under the ghostly glow of flickering torches. Shadows danced wildly on the walls as thunder roared outside, shaking ancient foundations. Tears streaked down Elara's face while Cedric brandished a glinting sword with trembling hands. In a heart-wrenching confrontation, secrets of betrayal and lost love unfurled like sinister specters around them. As their voices rose in anguished discord, spectral figures emerged from hidden corners—victims of past decisions demanding retribution. Lightning flashed through stained glass windows, illuminating their tormented faces in surreal clarity. At that moment of climax, time suspended; Castle Watchtower seemingly inhaled with anticipation. Elara spoke in fractured whispers about seeing lights dancing ominously at deserted chambers—a sign she interpreted as her unavoidable doom. Each word felt like shards piercing Cedric’s conscience—a haunting echo fueling irrevocable choices made long ago but tearing apart everything they shared
\\ \hline
\end{tabular}
\caption{The second example of Text-Only story element generation. They are less creative than IG's results.}
\end{table}

\begin{table}[h!]
\centering
\begin{tabular}{|p{15cm}|}
\hline
\textbf{DOC partial example 2} \\ \hline
Hi, I am Elena. As shadows of my past echo in their confrontation, I am drawn to witness. Memories and guilt swirl, urging me to guide them toward reconciliation. In the haunting silence, I offer understanding and hope for redemption in their choices. Elara’s voice trembles as she recounts her forbidden love Elara’s voice trembles as she recounts her forbidden love, each word shedding a light on the darkness that has enveloped her heart for years. I step closer, my breath mingling with the chill in the grand hall, creating misty wisps that dissolve into the ghostly atmosphere. Cedric's eyes burn with a mixture of anger and sorrow, the glint of his sword casting fleeting reflections across the floor. ``Why didn't you tell me?'' Cedric's voice cracks, revealing a depth of pain masked by years of silence. Elara's eyes meet his, tears shimmering like fractured glass. ``I couldn't bear to break you, Cedric,'' she whispers, her voice barely audible yet heavy with regret. ``Our family demanded so much from us—duty above all else. I sacrificed my happiness to preserve the legacy... but it wasn't enough.'' I press my hand against the cold stone wall, feeling its rough texture beneath my fingertips. The castle seems to breathe, each inhalation echoing its ancient sorrow. I step forward, each footfall muffled by the thick fog that clings stubbornly to the ground. The iron gate behind me creaks ominously as it closes, shutting me in with my past. The grand hall looms ahead, its once-majestic doors now weathered by time and neglect. As I approach, the whispers grow louder, a symphony of grief and accusation that chills me to my core. The fog swirls around me, revealing glimpses of familiar faces—my father’s stern visage, my mother 's sorrowful eyes, my siblings' silent, judgmental stares. Each phantom is a reminder of the weighted history I carry within me. I push open the grand doors and they groan in protest, as though resisting my intrusion into what remains of their solemn dignity. Inside, the hall is a chasm of ghostly memories and lingering regret. The torches flicker as if whispering secrets yet to be told, casting wavering shadows on the timeworn tapestries that tell stories of honor now tarnished by my deeds. Elara’s voice rises above the eerie silence, filled with anguish and accusation. Her eyes, filled with unshed tears, reflect the torment of betrayed hopes. ``How could you, Cedric? You knew everything,'' she cries out, her voice cracking like the lightning that splits the ominous sky outside. Cedric's sword trembles in his grip, mirroring his internal conflict. ``Elara, you don't understand,'' he responds, his voice strained under the weight of decades-old secrets. ``I did what I thought was right for the family.'' As their words clash, the ghosts within Castle Watchtower seem to stir more violently, their ethereal forms flickering in and out of existence along with every accusation hurled. The specters' presence grows stronger with every word exchanged between Elara and Cedric. I take a hesitant step forward, feeling the bitter cold of the castle floor seep through my boots. The echoes of their voices reverberate in my ears, mingling with the whispers of my own past transgressions. ``I must guide them,'' I whisper to myself, steeling my resolve. The flickering torches cast eerie shadows on the walls as I approach Elara and Cedric, their faces twisted in anguish and fury. Cedric's sword catches the light, reflecting the storm raging within and outside these ancient walls. ``Stop '' I cry out, my voice trembling but firm. Cedric's eyes snap to me, confusion and anger warring in his gaze. Elara's tear-streaked face turns toward me, her expression one of desperate curiosity and pain. ``Who are you?'' Cedric demands, his grip on the sword tightening. 
\\ \hline
\end{tabular}
\caption{This is part of the second final result of the DOC. The full generated result (about 2.5k) is too long, so we only report a portion of the full sample. You can see the full story in \url{https://www.2024cci.p-e.kr/4o/vsDOC/view.html?n=2}}
\end{table}

\begin{table}[h!]
\centering
\begin{tabular}{|p{15cm}|}
\hline
\textbf{Text-only imagined character example 3} \\ \hline
Elliot : 42 years old. a tall, slender man. I have shaggy dark hair and weathered skin.

\\ \hline
\textbf{Text-only imagined background example 3} \\ \hline
 a storm-ravaged coastal town, shrouded in thick fog and battered by relentless waves. The few remaining residents cling to their memories, as whispers of old legends echo through deserted streets. Broken lampposts flicker erratically, casting eerie shadows on crumbling buildings. The salty sea air carries distant cries of seagulls, mingling with the sound of crashing tides. Amidst the chaos, a mysterious lighthouse stands solemnly atop a cliff, its light cutting through the gloom like
\\ \hline
\textbf{Text-only imagined main plot example 3} \\ \hline
A fierce storm rages as the townspeople gather at the cliff's edge, their faces etched with despair and hope. The lighthouse beam flickers erratically, mirroring their inner turmoil. Waves crash violently against the rocks, sending up salty sprays that drench them. In the throes of emotional turmoil, a pivotal figure stands firmly against the wind, holding onto a cherished token of lost love. Tears mingle with rain as they confront their haunting memories and unspoken truths. Just then, a colossal wave rises ominously behind them—its deafening roar silences all murmurs. At once beautiful and terrifying, it looms over everything before crashing down with relentless force—a moment frozen in time as everything becomes chaos or clarity for those present.
\\ \hline
\end{tabular}
\caption{The third example of Text-Only story element generation.}
\end{table}

\begin{table}[h!]
\centering
\begin{tabular}{|p{15cm}|}
\hline
\textbf{DOC partial example 3} \\ \hline
 As I stand at the edge of the cliff, the wind howls around me like a grieving spirit, tugging at my worn leather jacket. The sea below churns with a fury that reflects the storm within me. My scarred cheek stings with each cold raindrop , a stark reminder of battles past. As the storm rages, I stand resolute at the cliff's edge, clutching my scar as a reminder of battles past. The rain mingles with my tears, blurring the line between sky and sea. My fingers curl tightly around a weathered locket—inside it, a faded picture of her smile The wind howls like a mournful choir, and the cold bite of salt stings my weathered face. Lightning streaks across the sky, momentarily revealing the faces of those gathered—their eyes wide with shared dread and silent prayers. The lighthouse beam flickers frantically, casting ghostly shadows that dance upon the jagged rocks below. I tighten my grip on the silver locket, its cold metal pressing into my palm, a tangible link to a love long lost yet never forgotten. Lucy stands among the crowd, her pragmatic demeanor betraying an undercurrent of worry. She catches my eye briefly and gives a solemn nod— —a small gesture of support amidst the chaos. Her presence is both comforting and painful, a reminder of the family that’s been left behind by storms, both literal and figurative. As I look out over the tumultuous sea, the waves rise higher and more ferociously than I’ve ever seen them. The wind lashes against my face, stinging my eyes and making it hard to keep them open. The distant sound of the lighthouse horn cuts through the cacophony, its eerie wail resonating with the haunting memories that flood back to me—my friend’s desperate cries swallowed by the tempest. The townspeople shift uneasily as the wind continues its relentless assault. Faces drenched, their expressions mirror my own—an uneasy blend of fear and hope. I tighten my grip on the silver locket, feeling its cold metal press into my palm, a tether to a past that still has its claws in me. The lighthouse beam sweeps over the tumultuous scene again, its light catching the faces of those gathered. I spot a fisherman, rugged and stoic, staring out at the horizon as if searching for answers that elude him. To my right, a mother clutches her child protectively to her chest, casting anxious glances toward the furious sea as if silently pleading for the storm's mercy. The child's wide eyes, full of both fear and curiosity, remind me of my younger self—innocent yet marked by uncertainty. The gusts grow fiercer, making it hard to stand upright. My leather jacket flaps violently in the wind, and I can feel the cold seeping through every thread. The locket in my hand feels like an anchor, weighing down the chaos swirling within me. My heart hammers against my ribcage as I lock eyes with her—my lost love’s sister, Pauline. Her piercing gaze cuts through the storm, searching for answers that have eluded us both for years. She steps closer, her face etched with a mix of sorrow and determination. The storm's force tugs at her coat, but she stands tall, undeterred. I feel the relentless wind pushing me nearer to the edge, each gust stoking the fires of my guilt and regret. The murmur of the townspeople fades into background noise, drowned by the howling wind and crashing waves. My vision blurs momentarily from the mix of rain and tears, but Pauline's determined stance remains clear. She raises her hand towards me—a bridge over years of silence and unresolved pain. I can't discern her words amidst the thunder, but the intent is unmistakable: reconciliation. I take a step toward her, feeling every grain of sand under my boots as if it carried the weight of my past. In this storm's unrelenting grip, fragments of old memories and regrets surge into clarity. I see my friend's face, twisted in fear as the waves consumed him, and my own paralyzing terror that sealed his fate. (...)
\\ \hline
\end{tabular}
\caption{This is part of the third final result of the DOC. The full generated result (about 2.5k) is too long, so we only report a portion of the full sample. You can see the full story in \url{https://www.2024cci.p-e.kr/4o/vsDOC/view.html?n=1}}
\end{table}

\twocolumn

\section{Example of CCI}
\label{sec:appendixh}
We report the three final results of CCI-4o, which were used in human and LLM evaluations. Therefore, these are not cherry-picked results. First, we show the characters, backgrounds, and climaxes generated by IG, along with the actual images. IG's qualitative performance is proven by its overall high diversity and specificity, especially in terms of the character's imagination.

Subsequently, the Specification stage further expands the characters and main plot. These detailed elements are used as prompts in the Plan and Draft stages. In particular, MW demonstrates qualitative performance by inserting personal traits in detailed character descriptions into the story. As a result, compared to the output of DOC, MW clearly provides more frequent and detailed clues about the protagonist. Thus, our core contributions, IG and MW, qualitatively enhance the specific imagination and add novelty to the story.
\onecolumn

\begin{table}[h!]
\centering
\begin{tabular}{|p{15cm}|}
\hline
\textbf{CCI character example 1} \\ \hline
\begin{minipage}{0.45\textwidth}
    \vspace{3mm}
    \centering 
    \includegraphics[width=0.9\textwidth]{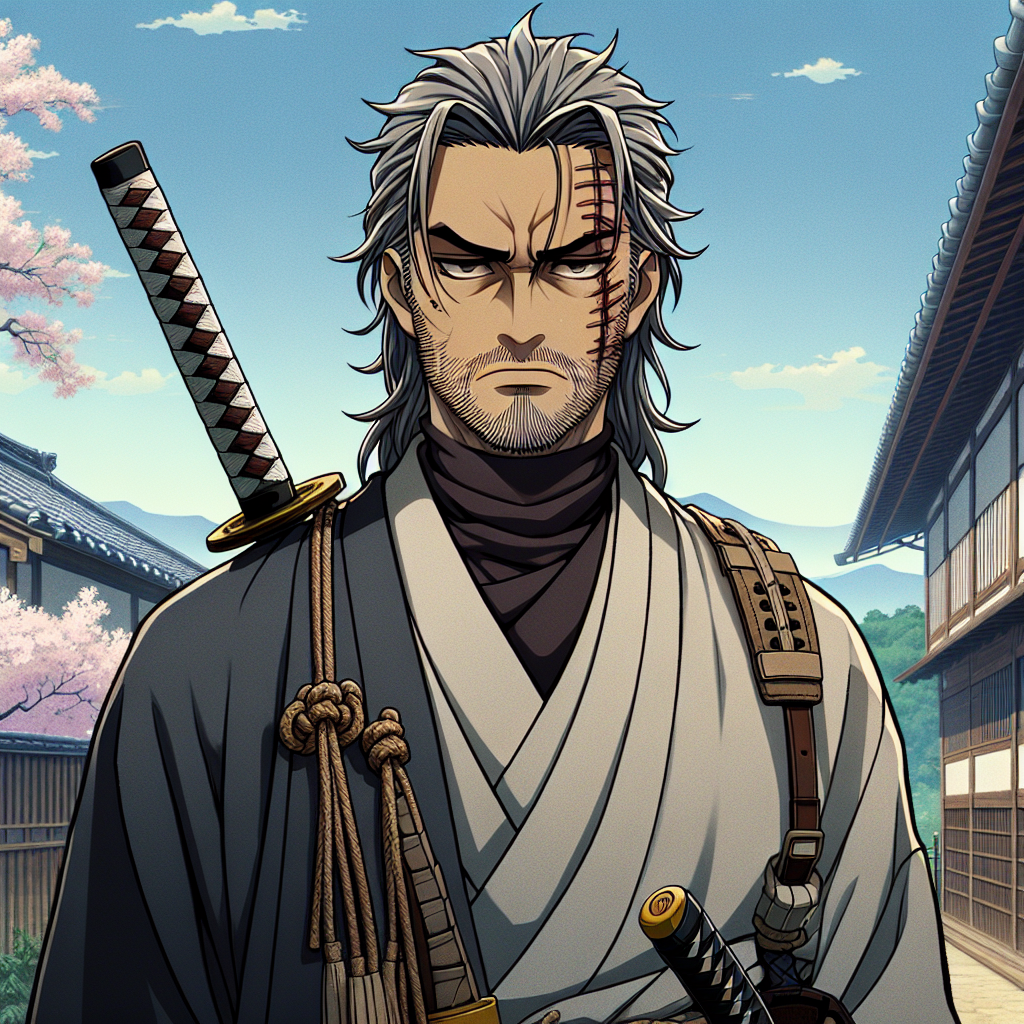} 
  \end{minipage}%
  \hspace{3mm}
  \begin{minipage}{0.45\textwidth}
  \vspace{3mm}
 Hiro: a rugged warrior with silver hair and a scarred face. He carries two swords and wears traditional samurai robes. His intense eyes reflect a fierce determination. He stands in a serene village with blooming cherry blossoms. Rope accents his attire, adding to his stern, disciplined appearance. Beneath his stern exterior, wisdom and experience resonate. The tranquil background contrasts his tough demeanor. His presence commands respect and caution.

  \end{minipage}

\\ \hline
\textbf{CCI background example 1} \\ \hline

\begin{minipage}{0.45\textwidth}
    \vspace{3mm}
    \centering
    \includegraphics[width=0.9\textwidth]{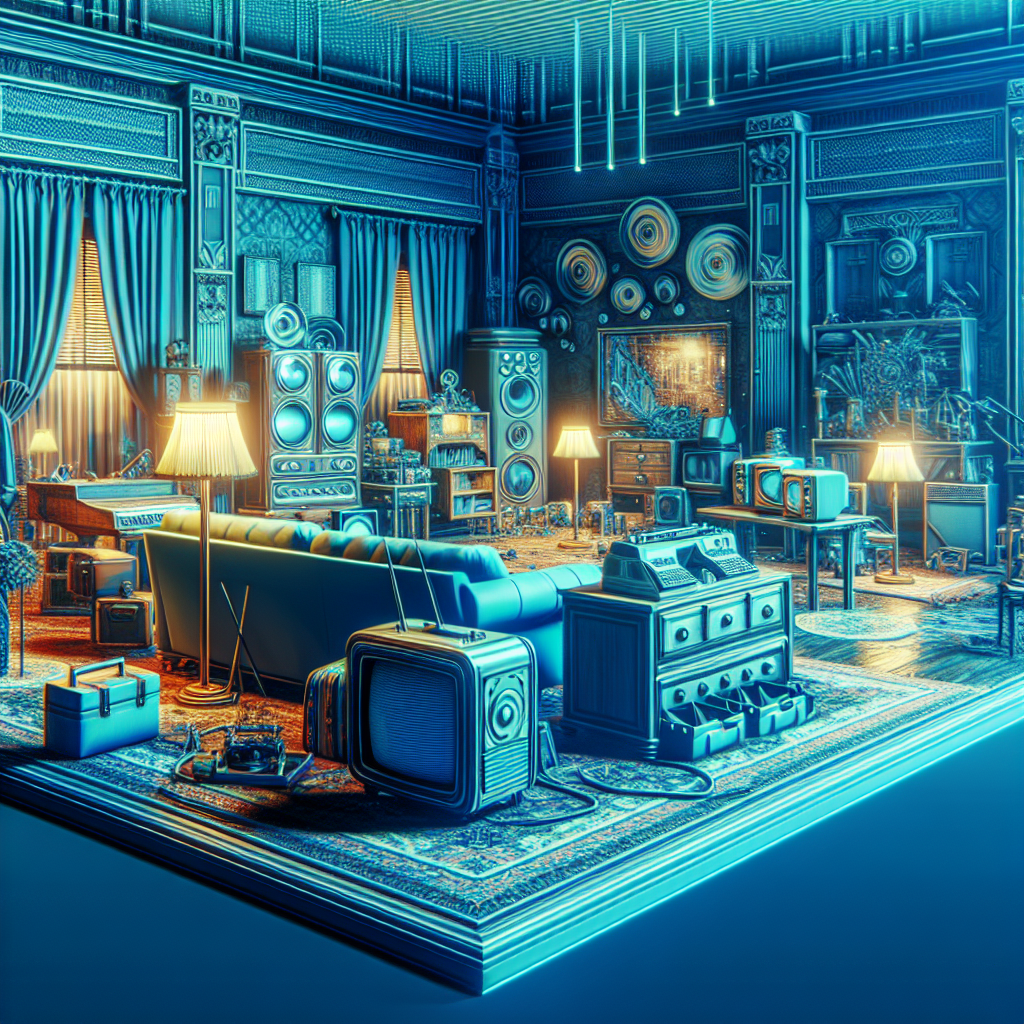}
  \end{minipage}%
  \hspace{3mm}
  \begin{minipage}{0.45\textwidth}
  \vspace{3mm}
The room is filled with various vintage electronics, including old TVs and speakers. Ornate curtains and rugs add a sense of opulence. The dim lighting from multiple lamps creates a cozy atmosphere. Wooden furniture and intricate woodwork adorn the walls. The setting feels like a blend of a retro technology hub and a luxurious library.

  \end{minipage}

\\ \hline
\textbf{CCI main plot example 1} \\ \hline

\begin{minipage}{0.45\textwidth}
    \vspace{3mm}
    \centering
    \includegraphics[width=0.9\textwidth]{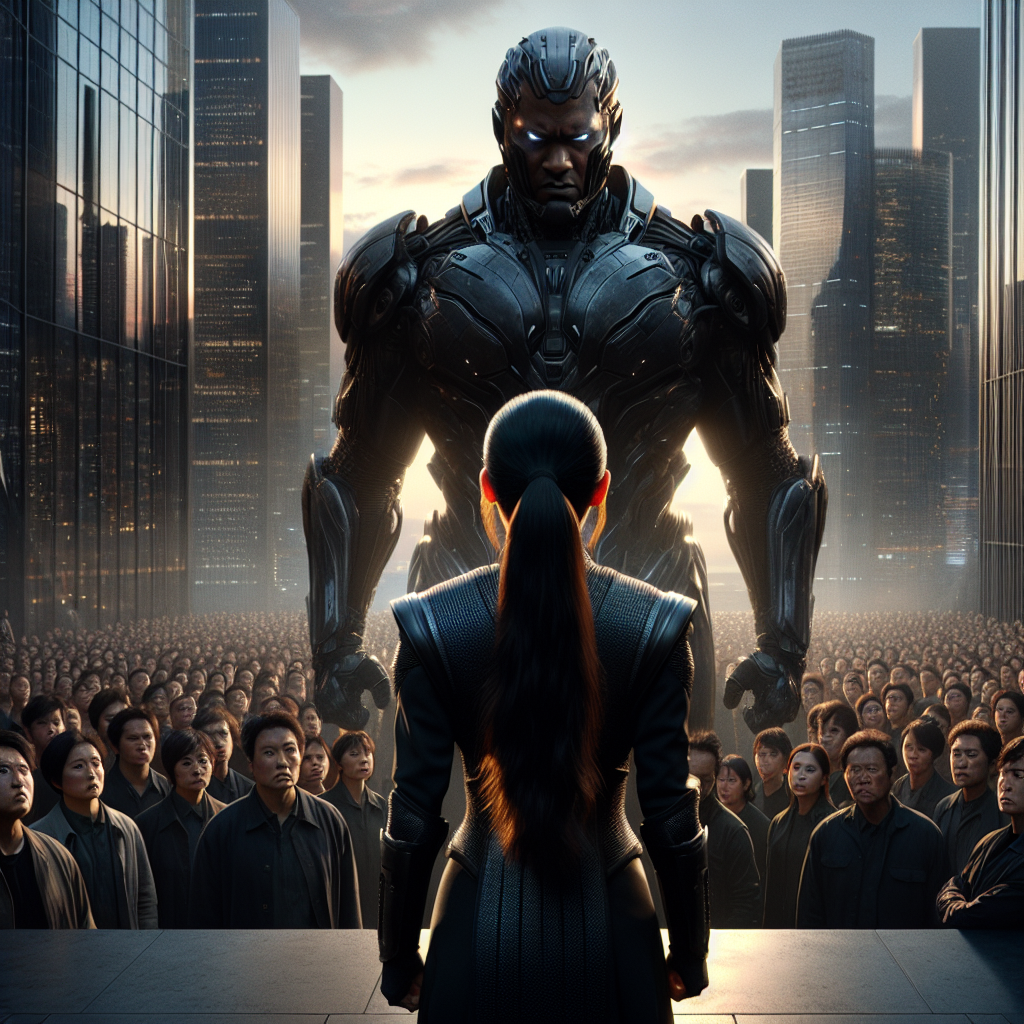}
    \vspace{3mm}
  \end{minipage}%
  \hspace{3mm}
  \begin{minipage}{0.45\textwidth}
  \vspace{3mm}
In a futuristic city, a lone figure confronts a towering armored warrior. The crowd watches in tense silence. The figure stands defiant, ready to challenge the imposing threat. The atmosphere is charged with anticipation and fear.
\vspace{3mm}
  \end{minipage}
\\ \hline
\end{tabular}
\caption{This is the first example of Image-guided story elements generation. They are more visually detailed than text-only story elements especially for the character. }
\end{table}

\begin{table}[h!]
\centering
\begin{tabular}{|p{15cm}|}
\hline
\textbf{CCI persona example 1} \\ \hline
\begin{enumerate}
\item \textbf{\textit{What is your dark secret?}} Beneath my stern exterior and unwavering determination lies a secret I've buried deep in the recesses of my heart. I was once part of a covert group that dismantled and sold stolen vintage electronics from wealthy homes. The intricate woodwork and ornate curtains in my room once belonged to a noble house I helped rob. My disciplined demeanor masks the haunting guilt that these treasures I cherish are trophies of my darkest deeds. It’s a burden that continuously reminds me of my tainted past, even as I stride
\item \textbf{\textit{What is your family environment like?}} My family environment is one of quiet strength and resilience. My father was a disciplined samurai, teaching me the way of the sword from a young age. My mother was wise, instilling in me a deep sense of honor and wisdom. Though our communication was sparse, every word carried weight. Our home was modest but rich with tradition. The presence of vintage electronics added a modern touch to our timeless setting, reflecting my father's fascination with both the past and the future. The wooden furniture, ornate curtains
\item \textbf{\textit{Describe your appearance in detail.}} In my self-portrait, I see a man whose silver hair flows just past his shoulders, streaked with the wisdom of years. A prominent scar runs diagonally across the left side of my rugged face. My intense eyes, dark with shadows of the past, exhibit a fierce determination. I carry dual swords at my sides, their hilts worn from countless battles. I wear traditional samurai robes, layered in shades of gray and white, signifying both simplicity and depth. The rope accents
\item \textbf{\textit{What is your specific way of speakings(tone of speech)?}} My tone of speech is grounded yet respectful, with a trace of inherent wisdom. My words are often concise and to the point. Here are some examples: 1. **Inspiring Others:** ``Strength lies within. Honor your path, and it will honor you in return.'' 2. **Offering Wisdom:** ``Patience is the samurai's greatest weapon. Rash decisions lead to regret.'' 3. **In Battle:** ``Stand firm. Our resolve will outlast the storm.'' 4. **Address
\item \textbf{\textit{What is your personality like?}} I, Hiro, embody conscientiousness (CON). Here are examples showcasing my personality: 1. **Precision in Battle**:   My careful planning and strategic thinking ensure triumph. Every move is calculated, reflecting my disciplined nature. 2. **Maintenance of Equipment**:   My swords are immaculately cared for, maintained to perfection. This dedication keeps them sharp and reliable. 3. **Commitment to Rituals**:   Performing daily meditation and training fosters my mental and physical well-being. These
\item \textbf{\textit{What are the most significant events you have experienced from childhood to now, primarily those that left a **trauma**? }} At a young age, I witnessed the brutal attack on our village by marauders. My family perished, and the trauma of their screams still haunts me. During my teenage years, I trained under a master who was stern yet wise. His sudden assassination left me feeling abandoned and vengeful. In my early twenties, I was forced into a war. The battlefield horrors and the loss of comrades left deep psychological scars. A mission to protect a noble family ended in failure. The entire household
\item \textbf{\textit{What is your very specific habitual behaviors?}} 1. **Sharpening Swords:** Daily, meticulously tending to his blades. 2. **Meditative Silence:** Spending mornings in deep meditation. 3. **Routine Inspections:** Conducting precise checks of his surroundings. 4. **Tea Ceremony:** Performing a traditional tea ceremony each evening. 5. **Strategic Reflection:** Analyzing battles and strategies during nightfall.
\end{enumerate}
\\ \hline
\end{tabular}
\caption{This is the result of character specification. We detail the character's internal and external attributes and use them as a persona during the draft. }
\end{table}

\begin{table}[h!]
\centering
\begin{tabular}{|p{15cm}|}
\hline
\textbf{CCI partial example 1} \\ \hline
I narrow my eyes, refusing to let his words seep into my resolve. ``Not this time,'' I reply, teeth gritted as I parry a powerful blow aimed at my shoulder. The force of his strike reverberates through my arms, nearly pushing me off balance. The ancient skyscraper groans under the strain of our battle, shards of glass crunching beneath our feet with each calculated step.

Lightning flashes across the sky once more, casting eerie shadows that dance around us. My swords clash against his massive \textbf{blade, sparks flying into the chaotic air. The wind howls, bending the tops of those towering structures as if bowing to the power of our confrontation.}

\textbf{I focus intently, channeling every ounce of my training and experience. The environment mirrors my internal state—turbulent yet controlled. His taunts might}  have broken a lesser man, but I hold firm. Each taunt fuels my resolve, pushing me to fight harder. He swings his colossal blade with lethal precision, yet I manage to parry and counterstrike in a deadly dance of steel.

One misstep from him—just one—and I seize the moment. My blade slices through the air with precision, striking true against his exposed side. The armored warrior stumbles, a guttural growl emanating from beneath his helmet. Seizing the advantage, I surge forward, relentless in my assault. Each strike and parry resonates with years of training and the weight of my past. \textit{With every clash of our swords, the echoes of my ancestors guide my hand. The warrior before me—though formidable—is not invincible. I can sense the hesitation behind his guard, the flicker of doubt in his eyes. In this crucible of combat, honor and skill are the true arbiters of victory. (...)}

\\ \hline
\end{tabular}
\caption{This is part of the first result of the CCI. The bold sentences describe the relationship between the protagonist and the enemy. The italicized sentences were meant to depict the protagonist speaking, but MW ultimately failed to do so. These are the result of persona injection by the Multi-writer model. You can see the full story at \url{https://www.2024cci.p-e.kr/CCI_Generations/4o/view.html?n=1}.}
\end{table}

\begin{table}[h!]
\centering
\begin{tabular}{|p{15cm}|}
\hline
\textbf{CCI character example 2} \\ \hline
\begin{minipage}{0.45\textwidth}
    \vspace{3mm}
    \centering
    \includegraphics[width=0.9\textwidth]{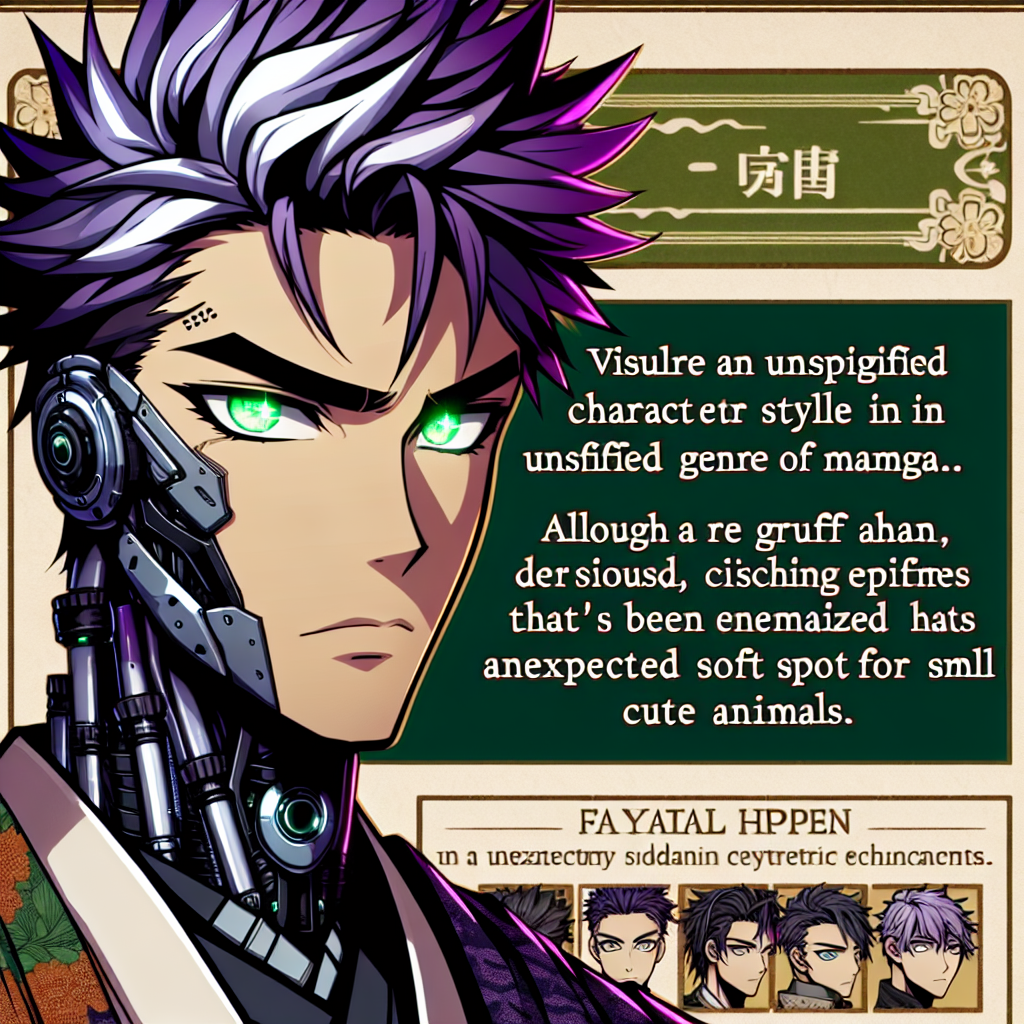} 
  \end{minipage}%
  \hspace{3mm}
  \begin{minipage}{0.45\textwidth}
  \vspace{3mm}
 Rex: a strong, stoic warrior. He has vibrant, spiky purple hair and glowing green eyes. His left side is mechanical. His gaze is intense and unyielding. He wears a traditional robe adorned with intricate patterns. A mysterious aura surrounds him. He is known for his unexpected soft spot for small, cute animals.
  \end{minipage}

\\ \hline
\textbf{CCI background example 2} \\ \hline

\begin{minipage}{0.45\textwidth}
    \vspace{3mm}
    \centering
    \includegraphics[width=0.9\textwidth]{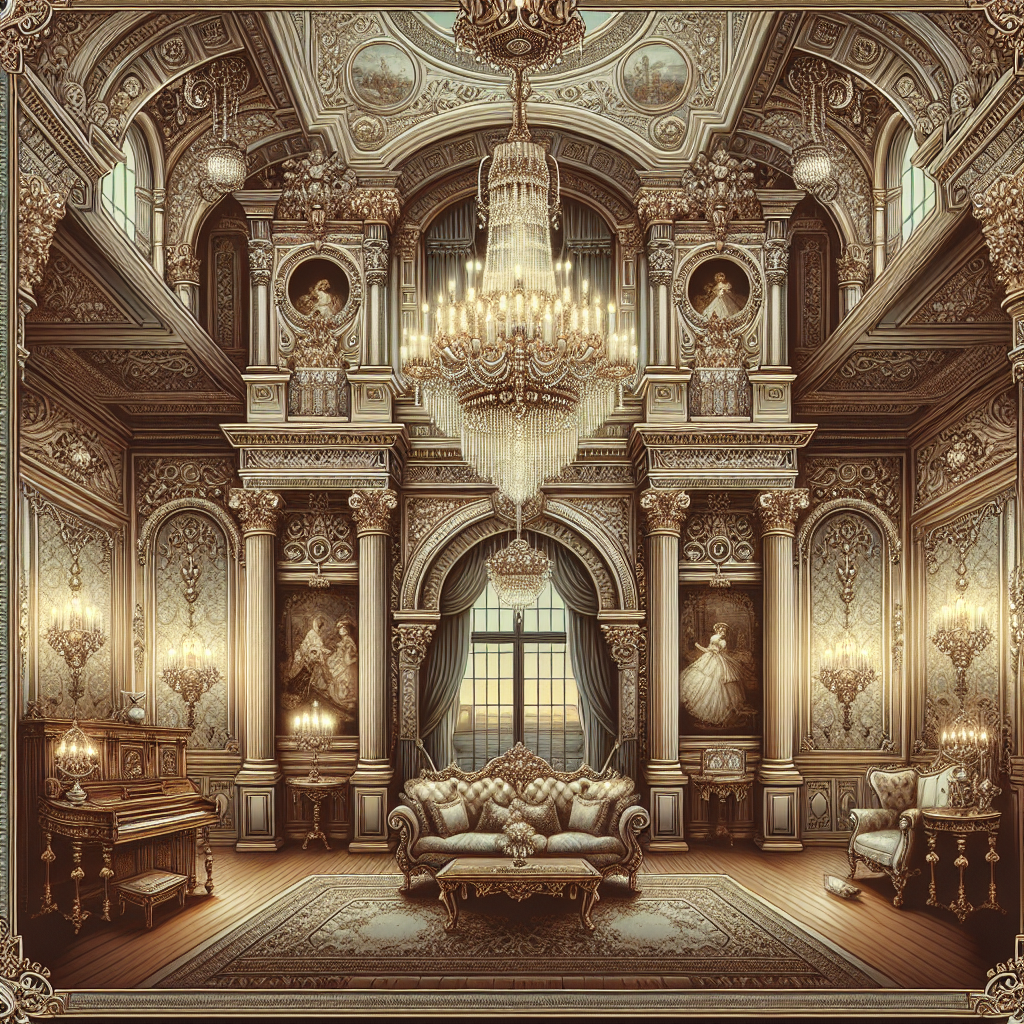}
  \end{minipage}%
  \hspace{3mm}
  \begin{minipage}{0.45\textwidth}
  \vspace{3mm}
The grand hall, adorned with intricate gold details, exudes opulence from every corner. Tall windows with velvet drapes frame a serene ocean view in the distance. Crystal chandeliers hang gracefully from the high ceiling, casting a warm glow. The room is furnished with luxurious, vintage-style couches, a piano, and ornate side tables. Timeless portraits and exquisite decor create an atmosphere of historic elegance and nobility.

  \end{minipage}

\\ \hline
\textbf{CCI main plot example 2} \\ \hline

\begin{minipage}{0.45\textwidth}
    \vspace{3mm}
    \centering
    \includegraphics[width=0.9\textwidth]{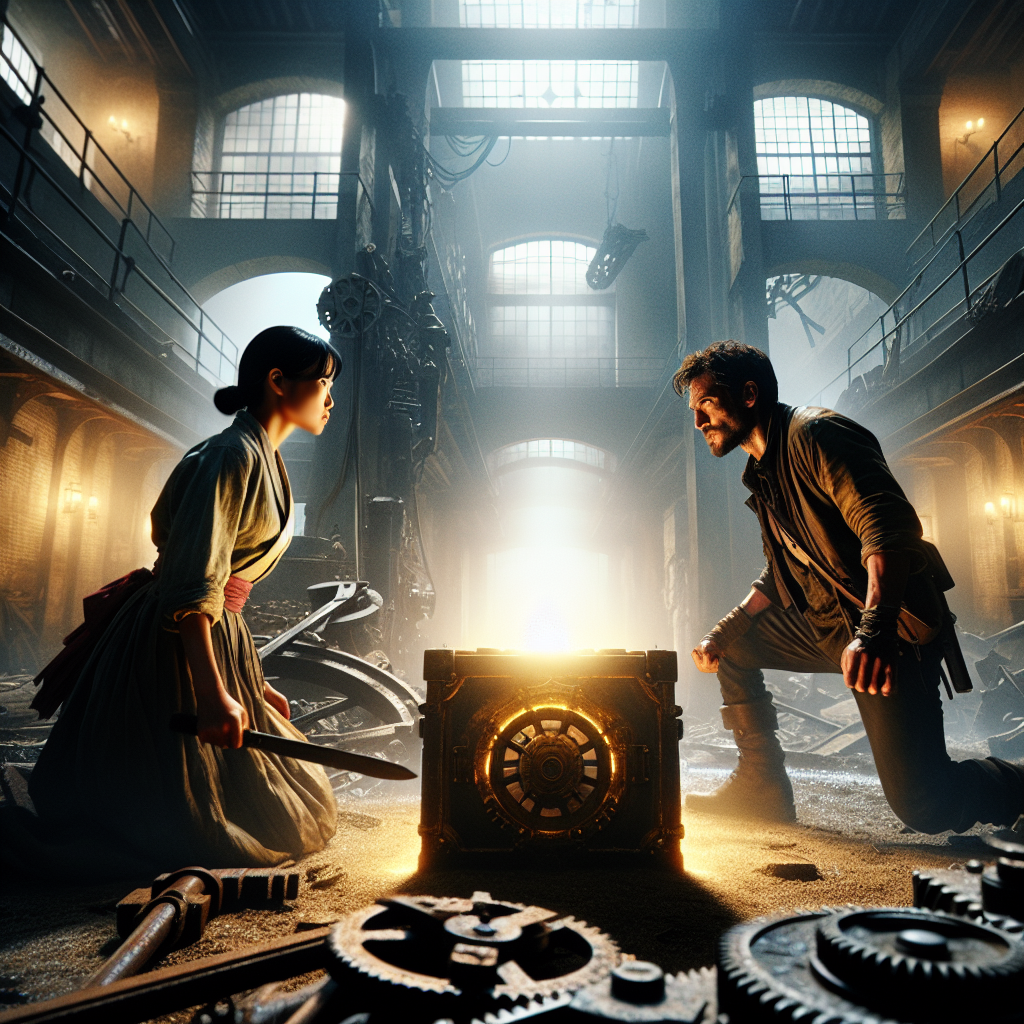}
    \vspace{3mm}
  \end{minipage}%
  \hspace{3mm}
  \begin{minipage}{0.45\textwidth}
  \vspace{3mm}
In the dimly lit industrial hall, two determined individuals confront each other over an ancient, glowing artifact. The woman brandishes a knife, while the man stands ready with clenched fists. Mechanical gears and tools lie scattered around them, hinting at a past struggle. Sunlight streaks through broken windows, illuminating their tense standoff. The artifact hums with energy, poised to reveal its secrets. Their eyes lock, each silently daring the other to make the first move. The stakes are high, and neither is willing to back down.
\vspace{3mm}
  \end{minipage}
\\ \hline
\end{tabular}
\caption{This is the second example of Image-guided story element generation. }
\end{table}

\begin{table}[h!]
\centering
\begin{tabular}{|p{15cm}|}
\hline
\textbf{CCI persona example 2} \\ \hline
\begin{enumerate}
\item \textbf{\textit{What is your dark secret? }} My dark secret is that my mechanical side is filled with memories of those I failed to protect. Each metallic piece tells a tragic story. These guilt-ridden memories haunt me in silence, beyond my stoic exterior.
\item \textbf{\textit{What is your family environment like?}}In my family, warmth and discipline coexist. My father, a stern and honorable figure, passed down traditions and wisdom. My mother, gentle yet strong, offered nurturing love and taught compassion. My siblings and I, though raised with high expectations, found solace in each other. Our home, filled with respect and unity, was a sanctuary amidst the grandeur of our surroundings. The grand hall's opulence mirrored our noble lineage, but it was our bond that truly defined our strength. Amidst the historical
\item \textbf{\textit{Describe your appearance in detail. }} I have vibrant, spiky purple hair that stands out prominently. My eyes are glowing green, with an intense and unyielding gaze. On the left side, my face and body are partially mechanical, revealing intricate metal components and circuitry. I wear a traditional robe adorned with intricate patterns, combining ancient aesthetics with a modern twist. My presence is stoic and commanding, exuding a strong aura of mystery and strength. Despite my formidable appearance, I have a soft spot for small, cute animals,
\item \textbf{\textit{What is your specific way of speakings(tone of speech)?}} My way of speaking is firm and deliberate, every word chosen with precision. I speak with gravitas, ensuring that each sentence carries weight. Example: ``I stand by my choices, unwavering and resolute. Your words cannot shake my resolve.'' When it comes to guiding or instructing others, my tone remains authoritative yet respectful. Example:``Follow the path set before you, but never forget to listen to your heart. It will guide you truer than any map.'' Despite my stoic nature,
\item \textbf{\textit{What is your personality like?}} Based on my characteristics and life experiences in the grand hall, I exhibit conscientiousness (CON). I show diligence and precision in everything I do, whether it's maintaining the intricate patterns of my robe or ensuring the mechanical part of me operates seamlessly. For example, I systematically train every day to maintain my warrior skills, never missing a session. Additionally, despite my intense and stoic demeanor, I take great care in nurturing small, cute animals, often staying up late to ensure they’re safe and comfortable.
\item \textbf{\textit{WWhat are the most significant events you have experienced from childhood to now, primarily those that left a **trauma**? }} One of the earliest memories is losing my parents to a violent conflict. Raised by my grandfather, he was strict but loving. The day I lost my family estate in a fire remains etched in my mind. Each memory reduced to ashes. Despite my stern exterior, rescuing a stray kitten was a moment of pure, unguarded emotion. An injury in battle led to the loss of my arm. Adapting to my mechanical side was painfully challenging. I sheltered a hurt bird once. It reminded
\item \textbf{\textit{What is `your' very specific habitual behaviors?}} 1. **Daily Meditation:** Every dawn, Rex meditates by the grand windows, gazing at the ocean to center his thoughts. 2. **Mechanical Maintenance:** He routinely polishes and adjusts his mechanical parts at night to ensure that they are in perfect condition. 3. **Animal Visits:** Rex has a hidden corner in the grand hall where he keeps treats and toys for the small animals he encounters. 4. **Pattern Drawing:** He often sketches new intricate patterns for his robe during quiet moments,
\end{enumerate}
\\ \hline
\end{tabular}
\caption{This is the result of character specification. We detail the character's internal and external attributes and use them as a persona during the draft. }
\end{table}

\begin{table}[h!]
\centering
\begin{tabular}{|p{15cm}|}
\hline
\textbf{CCI partial example 2} \\ \hline
``I've seen friends fall,'' I continue, my voice gaining strength from the depths of memories of battles past. Each word carries the weight of countless sacrifices. ``Their faces haunt me every night, their voices echoing my failures.''

Her grip on the artifact loosens slightly, though it still glows with an otherworldly light that seems to pulse in time with our shared sorrow. The stained-glass window  bathes the moonlit garden in kaleidoscopic hues, casting shimmering patterns on the ancient statues and serene fountains. I feel a gust of cool night air brush against my face, mingling with the warmth emanating from the artifact between our hands. It's as if time itself has paused, allowing us this singular moment \textbf{of vulnerability amidst the chaos of our world. My vibrant, spiky purple hair whispers softly in the breeze, caught between its wild defiance and the serene stillness of the night. My glowing green eyes capture and reflect the fragmented colors from the window, intensifying the fierce resolve that has seen me through countless hardships.}

 The woman’s eyes glisten in the pale moonlight, each tear catching the glow of the artifact. Her fingers tremble beneath mine, a mirror to my own trembling hand. We are like two pillars holding up a bridge of sorrow and hope, connected by this ancient relic that burns with our collective pain and yearning.  My gaze locks onto the artifact, its shimmering light pulsating in rhythm with the beating of my heart. I feel every emotion etched into its surface—the echoes of battles fought, lives lost, and victories turned hollow. The moment stretches into eternity as my hand closes around it, the warmth seeping through my mechanical fingers. The energy courses through me, igniting fragments of memories—each one a painful reminder, each one a promise not to fail again. I see the faces of my comrades: brave, loyal souls who fell because of choices I made. They stand before me in silent testimony.

The woman steps closer, her eyes searching mine , \textbf{their depths brimming with unspoken questions and silent understanding. Her presence is a gentle reminder of the compassion my mother instilled in me, telling stories of love and resilience. I draw a deep breath, the scent of the ocean wafting through the grand windows, its salty breeze grounding me in the present.}

  The energy courses through me, igniting fragments of memories—each one a painful reminder, each one a promise not to fail again. I see the faces of my comrades: brave, loyal souls who fell because of choices I made. They stand before me in silent testimony. (...)
\\ \hline
\end{tabular}
\caption{This is part of the second result of the CCI. The bold sentences describe the characteristic appearance of the protagonist and the protagonist's meditation habits. These are the result of persona injection by the Multi-writer model. You can see the full story at \url{https://www.2024cci.p-e.kr/CCI_Generations/4o/view.html?n=2}.}
\end{table}

\begin{table}[h!]
\centering
\begin{tabular}{|p{15cm}|}
\hline
\textbf{CCI character example 3} \\ \hline
\begin{minipage}{0.45\textwidth}\
    \vspace{3mm}
    \centering
    \includegraphics[width=0.9\textwidth]{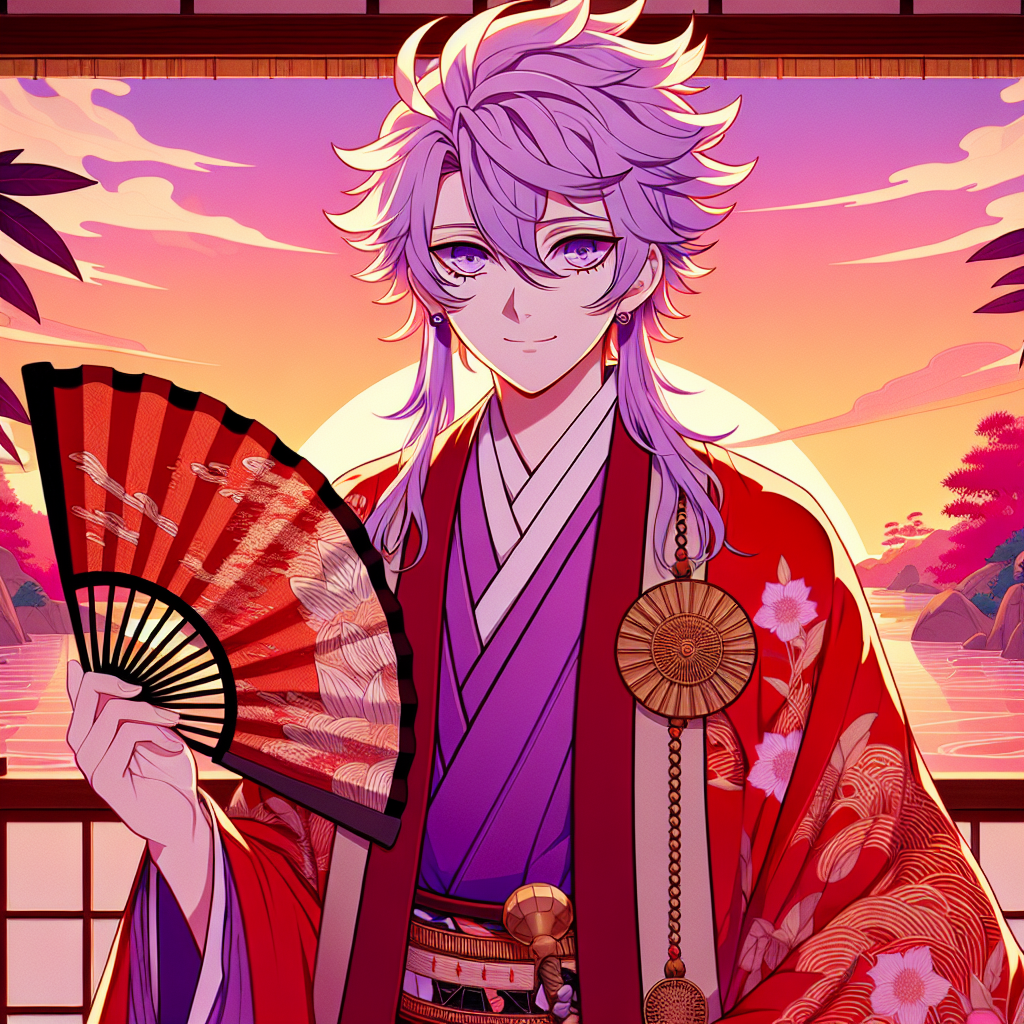} 
  \end{minipage}%
  \hspace{3mm}
  \begin{minipage}{0.45\textwidth}
  \vspace{3mm}
Yukio: a graceful figure. Flowing white hair. Wearing a vibrant red and purple kimono. Holds a delicate red fan. Soft, kind eyes. Golden sun adornments. Beautiful sunset backdrop. Poised and elegant demeanor.
  \end{minipage}

\\ \hline
\textbf{CCI background example 3} \\ \hline

\begin{minipage}{0.45\textwidth}
    \vspace{3mm}
    \centering
    \includegraphics[width=0.9\textwidth]{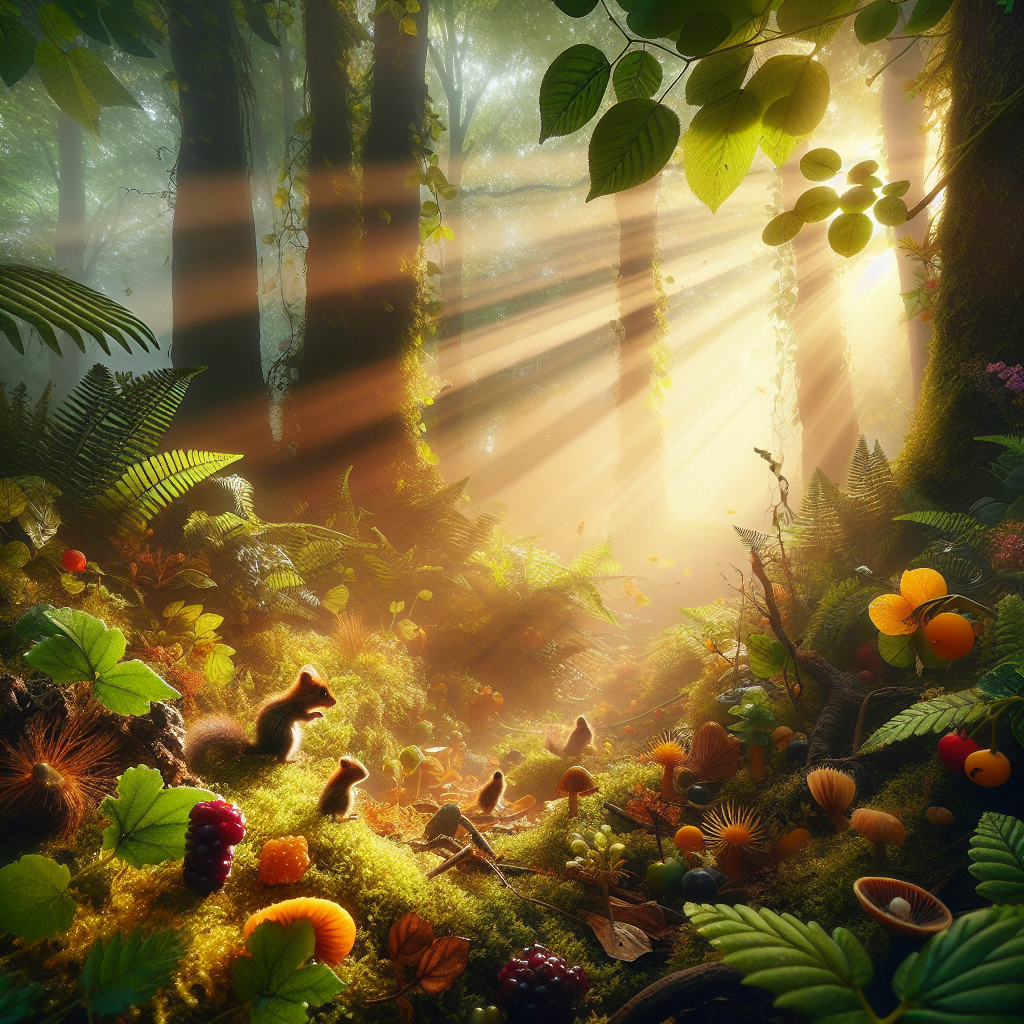}
  \end{minipage}%
  \hspace{3mm}
  \begin{minipage}{0.45\textwidth}
  \vspace{3mm}
Sunlight pierces through dense forest canopy, illuminating the lush greenery and vibrant flora. Ferns and wild mushrooms thrive in the moist environment, adding to the forest's richness. Small woodland creatures, like squirrels, move about, indicating a lively ecosystem. The air is fresh, filled with the scent of pine and earth. Rays of sunlight create a mystical ambiance, making the forest feel enchanted.
  \end{minipage}

\\ \hline
\textbf{CCI main plot example 3} \\ \hline

\begin{minipage}{0.45\textwidth}
    \vspace{3mm}
    \centering
    \includegraphics[width=0.9\textwidth]{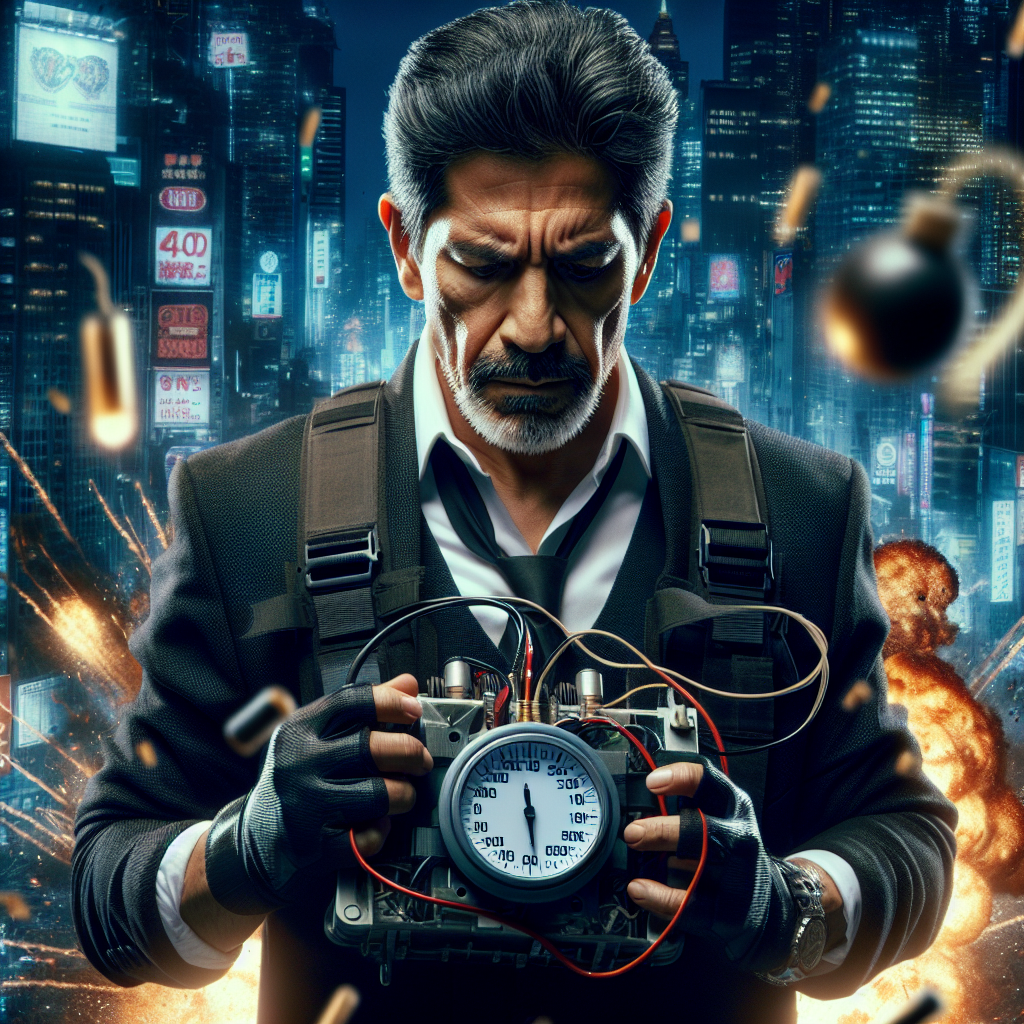}
    \vspace{1mm}
  \end{minipage}%
  \hspace{3mm}
  \begin{minipage}{0.45\textwidth}
    \vspace{3mm}
In a dense urban cityscape, a man defuses a bomb amidst chaos. Sweat drips down his face as time ticks away. Explosions and debris fly around. His determination might save countless lives. The city's fate rests on his shoulders in this tense, high-stakes moment.
\vspace{1mm}
  \end{minipage}
\\ \hline
\end{tabular}
\caption{This is the third example of Image-guided story elements generation.   }
\end{table}

\begin{table}[h!]
\centering
\begin{tabular}{|p{15cm}|}
\hline
\textbf{CCI persona example 3} \\ \hline
\begin{enumerate}
\item \textbf{\textit{What is your dark secret? }} My dark secret is that I once betrayed a dear friend to protect my family's honor. This guilt haunts me daily.
\item \textbf{\textit{What is your family environment like?}} My family environment is calm and harmonious. We live in a traditional home near the edge of the vibrant forest. The house is modest but well-kept, with sliding doors that open to the lush greenery outside. My parents are loving and hardworking, ensuring our home is filled with warmth and care. My mother tends a small garden where she grows herbs and flowers, while my father is a skilled artisan, crafting beautiful items from wood. I have a younger sister who is playful and curious, always exploring
\item \textbf{\textit{Describe your appearance in detail.}} For a more realistic and detailed description of my appearance: I stand tall with a graceful posture, my flowing white hair gently cascading around my shoulders. The strands are soft and silky, catching the light beautifully at every angle. My vibrant red and purple kimono is meticulously crafted from high-quality silk, adorned with intricate patterns that complement the natural world around me. In my hand, I hold a delicate red fan, painted with gold accents that shimmer subtly under the setting sun. My eyes carry a gentle kindness
\item \textbf{\textit{What is your specific way of speakings(tone of speech)?}} As Yukio, my tone of speech is calm, soothing, and poetic, often with a touch of melancholy. Here are some examples:1. *When greeting a traveler*: ``Welcome, weary soul, to this enchanted forest. May the whispering pines offer you solace.''2. *When describing the beauty around*: ``The golden rays of the sun gently kiss the emerald leaves, painting a serene tapestry of light and shadow.'' 3. *When advising someone*: ``In the quiet of 
\item \textbf{\textit{What is your personality like?}} My personality trait would be high in agreeableness (AGR). I am often seen helping small woodland creatures in my forest home, showing kindness and nurturing behavior. When friends visit, I offer them tea and a listening ear, reflecting my empathy and warmth. Additionally, I always strive to maintain harmony within my surroundings, often mediating disputes among the forest animals. A good example is when I recently helped two squirrels share their acorn stash, ensuring both were content. This
\item \textbf{\textit{What are the most significant events you have experienced from childhood to now, primarily those that left a **trauma**? }} Growing up, I faced the grief of losing my dearest friend to an unforgiving illness. In my teenage years, a devastating fire engulfed our home, leaving us with nothing but ashes and memories. A betrayal by someone I deeply trusted severed the bond, creating a chasm of mistrust within my heart. During a particularly harsh winter, I experienced starvation and cold, learning the depths of human resilience. In my adulthood, I witnessed the violence of war tearing apart families and
\item \textbf{\textit{What is ’your’ very specific habitual behaviors?}} 1. **Morning Meditation** - I start each day with a tranquil meditation by a serene forest stream, listening to nature's symphony. 2. **Gardening** - Tending to a small garden, I nurture delicate flowers and herbs, enjoying the peace it brings. 3. **Tea Ceremony** - Daily, I practice the art of the tea ceremony, cherishing each measured movement and the calm it instills. 4. **Scribing Poetry** - I write poetry inspired by the forest
\end{enumerate}
\\ \hline
\end{tabular}
\caption{This is the result of character specification. We detail the internal and external attributes of the character and use them as a persona during the draft. }
\end{table}

\begin{table}[h!]
\centering
\begin{tabular}{|p{15cm}|}
\hline
\textbf{CCI partial example 3} \\ \hline
 Hi, I am Yukio. Standing at the epicenter of the city's chaos, I feel a powerful blend of urgency and serenity thrumming within me. Dusk casts a golden hue over the cityscape, its turmoil softened by the beauty of the setting sun. In my hands, I hold both the power to destroy  and the power to save—represented physically by the intricate bomb before me and metaphorically by my choices that led to this moment. I draw a deep breath, envisioning each cherry blossom I left behind as a symbol of hope and renewal. 

The sound of hurried footsteps echoes in the distance, mingling with anxious murm \textbf{urs and a cacophony of voices. My mind drifts, almost involuntarily, back to the memories that shaped me into the man I am today.}

\textbf{I recall the anguished cries of my friend, his pain a cruel testament to the fragility of life. How I wanted to save him, to lend} him my strength and shield him from the shadows that consumed us both. Yet, as helpless as I felt then, the same sense of urgency propels me now to act with purpose. 

My fingers glide over the detonation mechanism, delicate yet deliberate. Each movement is precise, measured—the epitome of balance and control that  was nurtured through meditation and the teachings of the wise old hermit. The bomb's ticking synchronizes with my heartbeat, a rhythmic reminder of the stakes at hand. I breathe deeply, drawing in serenity from the memories of cherry blossoms and rippling brooks. (...)
\\ \hline
\end{tabular}
\caption{This is part of the third result of the CCI. The bold sentences describe the trauma of losing the protagonist's dearest friend, which is the result of persona injection by the Multi-writer model. You can see the full story at \url{https://www.2024cci.p-e.kr/CCI_Generations/4o/view.html?n=3}.}
\end{table}
\twocolumn



\section{Multimodal Interactive Story Generation}
\label{appendix:Multimodal Interactive Story Generation}

\begin{figure}
\centering
\includegraphics[width=0.5\textwidth]{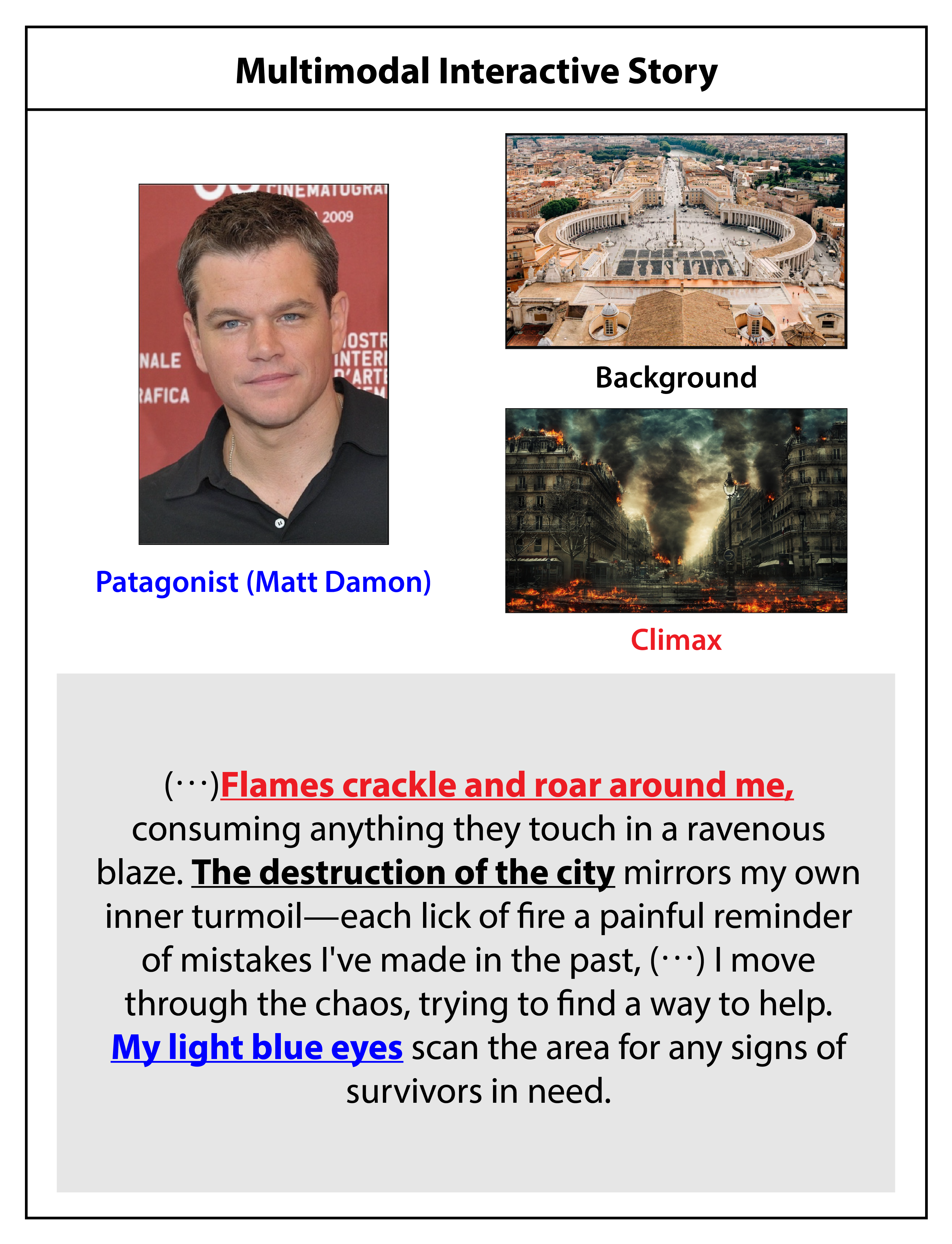}
\vspace{-3mm}
\caption{Example of Multimodal Interactive Story. We bold and underline the sentences that from the Protagonist, Background, and Climax images.}
\label{experiment_interactive}
\end{figure}

This is a example of the multimodal interactive story presented in section \ref{experiment_interactive}. We have bolded the sentences injected through MW's persona injection. You can see that the characteristics of the character, background, and climax images are well reflected, as well as other persona information such as behavioral habits, self-description, and specific speech patterns or tone. You can see other multimodal interactive stories at \url{https://www.2024cci.p-e.kr/Interactive/} 

\onecolumn

\begin{table}[h!]
\centering
\begin{tabular}{|p{15cm}|}
\hline
\textbf{Full Example of Multimodal Interactive Story} \\ \hline
The air is thick with smoke, making it hard to see or even breathe. Flames crackle and roar around me, consuming anything they touch in a ravenous blaze. The destruction of the city mirrors my own inner turmoil—each lick of fire a painful reminder of mistakes I've made in the past, particularly the cooking error that shattered my friendship with Ben. I can still see his face from that day, filled with hurt and betrayal. The thought makes my chest tighten. I move through the chaos, trying to find a way to help. My light blue eyes scan the area for any signs of survivors in need. The flames rise higher, turning the already reddish sky an ominous shade of orange. Memories of that fateful day in Ben's shop haunt me—flashes of fire, the smell of burnt food, his anguished voice echoing through time as vividly as the burning structures around me now . I can still hear Ben's frantic cry, ``Matt, what have you done?'' My heart races as if I'm back in that small kitchen, the air thick with smoke and the weight of my mistake. The irony isn't lost on me—fire seems to follow me, as if it senses my guilt. As I stagger through the square, the heat intensifies around me, licking at my skin and searing my lungs with every breath. Flames leap skyward, consuming everything in their path—ancient statues now veiled in a sinister shroud of fire. The once lively square is a terrifying vision of devastation. Swe Stagger through the square, the heat intensifies around me, licking at my skin and searing my lungs with every breath. Flames leap skyward, consuming everything in their path—ancient statues now veiled in a sinister shroud of fire. The once lively square is a terrifying vision of devastation. Sweat drips from my brow, stinging my eyes as I squint through the smoke. The air is thick and acrid, making each inhale a laborious effort. My heart pounds against my ribcage like a war drum, a rhythm of panic that refuses to be quelled. Suddenly, through the haze \textbf{I catch sight of a glimmer—a necklace I recognize. Instinctively, my hand reaches up to my own necklace for comfort, a brief but comforting gesture amid the chaos. The quick, small smile that typically precedes my words of reassurance falters, almost as if it's been swallowed by the turmoil surrounding me.} drips from my brow, stinging my eyes as I squint through the smoke. The air is thick and acrid, making each inhale a laborious effort. My heart pounds against my ribcage like a war drum, a rhythm of panic that refuses to be quelled. Suddenly, through the haze I catch sight of a glimmer—a necklace I recognize. Instinctively, my hand reaches up to my own necklace for comfort, a brief but soothing gesture amid the chaos. The quick, small smile that typically precedes my words of reassurance falters, almost as if it's been swallowed by the turmoil surrounding me. I push forward, driven by the glimpse of familiarity. The necklace belongs to my friend, who is somewhere in this inferno. The realization sends a jolt through me, galvanizing my weary limbs into action. With each step, I wade deeper into the scorching chaos, my mind torn between the destructive present and the haunting echoes of my past. The air is thick with acrid smoke, making every breath feel like a struggle against drowning in darkness. I can hardly see through the dense fog of ash and embers, but the urgency to find my friend keeps me moving. Each corner I turn seems to reveal more devastation— crumbling buildings, scorched remnants of once-lively shops, and eerie shadows cast by the flickering flames. The sights are eerily familiar, pulling me back to that awful day in my friend's shop when a simple mistake snowballed into a disaster. I try to shake off the memory as I navigate through the crumbling library, books scattered, fire-damaged walls, and suffocating smoke. Each step I take crunches on charred pages beneath my feet. My eyes sting from the acrid air as I search for survivors. Suddenly, I spot her—a woman pinned under a massive beam that has fallen from one of the collapsing bookshelves. Her face is contorted in pain, and her desperate eyes meet mine through the haze. I rush towards her, my heart pounding violently against my ribcage. Kneeling beside her, I realize that freeing her won't be easy. The beam is heavy and precariously balanced, and my hands still tremble from the earlier failed attempt. I clench my teeth, trying to steady myself, pushing away the suffocating thoughts of guilt and failure that threaten to overwhelm me. ``Hold on,'' I whisper, more to myself than to her, as I place my hands on the beam. The smoke swirls around me, making it difficult to see, let alone breathe. 
\\ \hline
\end{tabular}
\end{table}

\begin{table}[h!]
\centering
\begin{tabular}{|p{15cm}|}
\hline
The bell tower looms precariously above, its structure groaning with every passing second. I push against the beam with all my might. My muscles protest, but I can't afford to listen. ``Just... a little... more,'' \textbf{I mutter through gritted teeth, feeling the strain in every fiber of my body. My eyes, those same light blue eyes that have seen countless movie sets and red carpets, are now stinging from the smoke. The thin silver necklace around my neck feels heavier with each desperate breath I take. With a final push }, the beam shifts just enough for me to slip through. My heart pounds in my chest as I look around the burning cathedral aisle, trying to focus on any signs of life amid the chaos. The echoes of distant cries guide me forward, each step a battle against falling debris and scorching heat. I hear a faint wh imper over the crackling flames, and my eyes dart towards a small figure huddled near what used to be the altar. The child looks no older than six, their tear-streaked face illuminated by an eerie orange glow. They are clutching a ragged teddy bear tightly. Adrenaline surges through \textbf{ my veins as I make my way toward the child. Each movement must be precise; a wrong step could not only cost my life but theirs as well. My mind races, searching for the right words to offer comfort in such an unimaginable situation. }imper over the crackling flames, and my eyes dart towards a small figure huddled near what used to be the altar. The child looks no older than six, their tear-streaked face illuminated by an eerie orange glow. They are clutching a ragged teddy bear tightly. Adrenaline surges through my veins as I scramble over the smoldering wreckage, every muscle taut with urgency. My hands are burned and bloodied from clawing through the debris, but I barely register the pain; all I can think about is reaching that child. The smoke chokes me as I crawl, keeping my body low to the ground, my ears straining to hear the faint cries over the roar of the flames. Just ahead, I see her small figure huddled beneath the fallen beam, her eyes wide with terror. ``It's okay,'' I rasp out between coughs. ``I’m here to help you.'' My voice wavers with the effort, but I force myself to stay calm. The child’s chest rises and falls quickly, her breaths shallow and erratic. I take a moment to gauge the beam's weight before positioning my hands carefully. With a deep breath, I summon all the strength left in me and lift it off her small body. She lets out a small whimper as the pressure releases, and I quickly reach down to scoop her up into my arms. Her tiny fingers grasp at my shirt, and I can feel her trembling against me. Navigating through the blazing ruins of the square, I keep talking to her softly, trying to reassure her. ``It's okay, sweetheart. I’ve got you now. We’re going to be safe.'' Each word feels like a lifeline, not just for the child but for myself as well. As I step out from the choking smoke and chaotic inferno into the relative calm of a quiet park corner, I spot the lush greenery striving to stay alive amidst the soot and ash. It's surreal how this small oasis of nature remains untouched by the surrounding destruction. I set the little girl down gently on a patch of grass, her eyes wide with fear and gratitude. Breathing deeply for the first time in hours, I notice my friend covered in soot but mercifully unharmed, running towards me. Relief floods through my veins as we embrace tightly, the weight of past regrets pressing upon us both like the tangible soot that clings to our skin. ``I’m sorry,'' I choke out, the words almost lost amidst the background roar of chaos and
\\ \hline
\end{tabular}
\caption{This is the full result of multimodal interactive story generation. }
\end{table}

\end{document}